%% file: main.tex
\documentclass[pdflatex,sn-mathphys-ay,iicol]{sn-jnl}

\usepackage{graphicx}
\usepackage{multirow}
\usepackage{amsmath,amssymb,amsfonts}
\usepackage{booktabs}
\usepackage{colortbl}
\usepackage{bm}
\usepackage{xcolor}
\usepackage{textcomp}
\usepackage{listings}
\usepackage{url}
\providecommand{\burl}[1]{\url{#1}}
\newcounter{mainfigurecount}

\newcommand{\promptnote}[2]{%
    \vspace{0.5em}
    \noindent\textbf{#1}\quad#2\par\vspace{0.2em}
}
\newcommand{\prompttitle}[1]{%
    \vspace{0.6em}
    \noindent\textbf{#1}\par\vspace{0.15em}
}

\begin{document}

\title[From Transformation to Target State]{From Transformation to Target State: Rethinking Query Representation for Zero-Shot Composed Image Retrieval}

\author[1]{\fnm{Yihe} \sur{Zhao}}\email{23331020@bjtu.edu.cn}
\author*[1]{\fnm{Songhe} \sur{Feng}}\email{shfeng@bjtu.edu.cn}

\affil[1]{\orgdiv{School of Computer Science and Technology}, \orgname{Beijing Jiaotong University}, \orgaddress{\city{Beijing}, \postcode{100044}, \country{China}}}

\abstract{
Composed image retrieval (CIR) aims to retrieve a desired target image from a query consisting of a reference image and a modification text. This task exhibits an unusual representational asymmetry: the modification text specifies a transition from the reference state, whereas retrieval candidates depict completed target states. This creates a representation mismatch for zero-shot methods that query pretrained vision-language spaces directly with transformation-oriented language. We study this mismatch and reformulate zero-shot composed image retrieval as target-state reconstruction followed by retrieval. We instantiate this formulation with \textbf{ASAP-CIR}, a training-free framework that reconstructs a static target representation using a frozen multimodal large language model (MLLM). The representation combines multiple holistic descriptions with a variable set of importance-weighted atomic semantics, thereby preserving both overall target identity and fine-grained visual constraints. Retrieval then integrates holistic state alignment, atomic constraint grounding, and calibrated target-state evidence aggregation. A controlled text-only diagnostic shows that target-side static query formulations achieve more reliable retrieval than dynamic composed query formulations, particularly when source-state semantics must be suppressed or transformed. Experiments on FashionIQ, CIRR, and CIRCO further characterize the effectiveness and limitations of this representation principle, with the clearest gains on the multi-target CIRCO benchmark. These results show that how composed intent is represented before retrieval is a consequential design choice, distinct from the choice of retrieval backbone itself.
}

\keywords{Multimodal learning, composed image retrieval, pretrained vision-language models, zero-shot composed image retrieval}

\maketitle
\section{Introduction}
\label{sec:introduction}

Composed image retrieval (CIR) enables users to search for a desired image through a visual reference and natural-language feedback. Unlike conventional text-to-image retrieval, however, CIR does not directly provide a description of the target. It provides evidence about a reference state together with an instruction that specifies how this state should change. The retrieval system must infer the completed target state before it can compare the query with candidate images. This capability supports interactive product search, fashion recommendation, visual editing assistance, and open-domain image search, but also makes CIR fundamentally different from ordinary image-text matching.

Existing CIR methods can be roughly grouped into supervised and zero-shot paradigms. Supervised methods learn image-text composition from annotated triplets and can achieve strong performance when sufficient training data are available \citep{Vo_2019_CVPR,Baldrati_2022_CVPR,Han_2023_CVPR,Song_2024_CVPR}. Collecting such triplets is expensive, however, and models trained on fixed datasets may not generalize to new domains or unseen user intents. Zero-shot CIR (ZS-CIR) instead reuses pretrained vision-language models (VLMs), especially Contrastive Language-Image Pre-training (CLIP), or employs large language models (LLMs) and multimodal large language models (MLLMs) to infer target-related text \citep{Saito_2023_CVPR,Baldrati_2023_ICCV,karthik2024visionbylanguage,10.1145/3626772.3657740,Tang_2025_OSrCIR}. Although this removes task-specific training, it exposes a more fundamental issue: the input semantics describe a transition, while the retrieval space is organized around observable visual states.

We refer to this issue as the \emph{representation mismatch} in ZS-CIR. A modification such as ``remove the person'', ``make the bottle orange'', or ``add a wooden log'' denotes an operation rather than a completed visual state. Directly encoding such language may preserve removed entities, superseded attributes, or editing predicates that should not be visible in the target. The difficulty is therefore not merely to compose an image and a sentence; it is to represent the intended target state in a pretrained image-text space that was not explicitly trained to execute state transitions.

Our central insight is to reconstruct the target state before retrieval instead of asking a pretrained VLM to perform an implicit state transition inside its embedding space. Target-state reconstruction canonicalizes diverse dynamic instructions into static descriptions of what should be visible. Yet one caption is not sufficient for every composed query. The intended state has complementary holistic and atomic structure: holistic descriptions preserve target identity and scene coherence, whereas atomic semantics retain changed attributes, counts, relations, and newly introduced objects as explicit visual constraints. We therefore represent the reconstructed state as a static target representation containing both levels. MLLM reasoning remains restricted to the query side, while candidate features can be precomputed for efficient retrieval.

\begin{figure}[t]
    \centering
    \includegraphics[width=\linewidth]{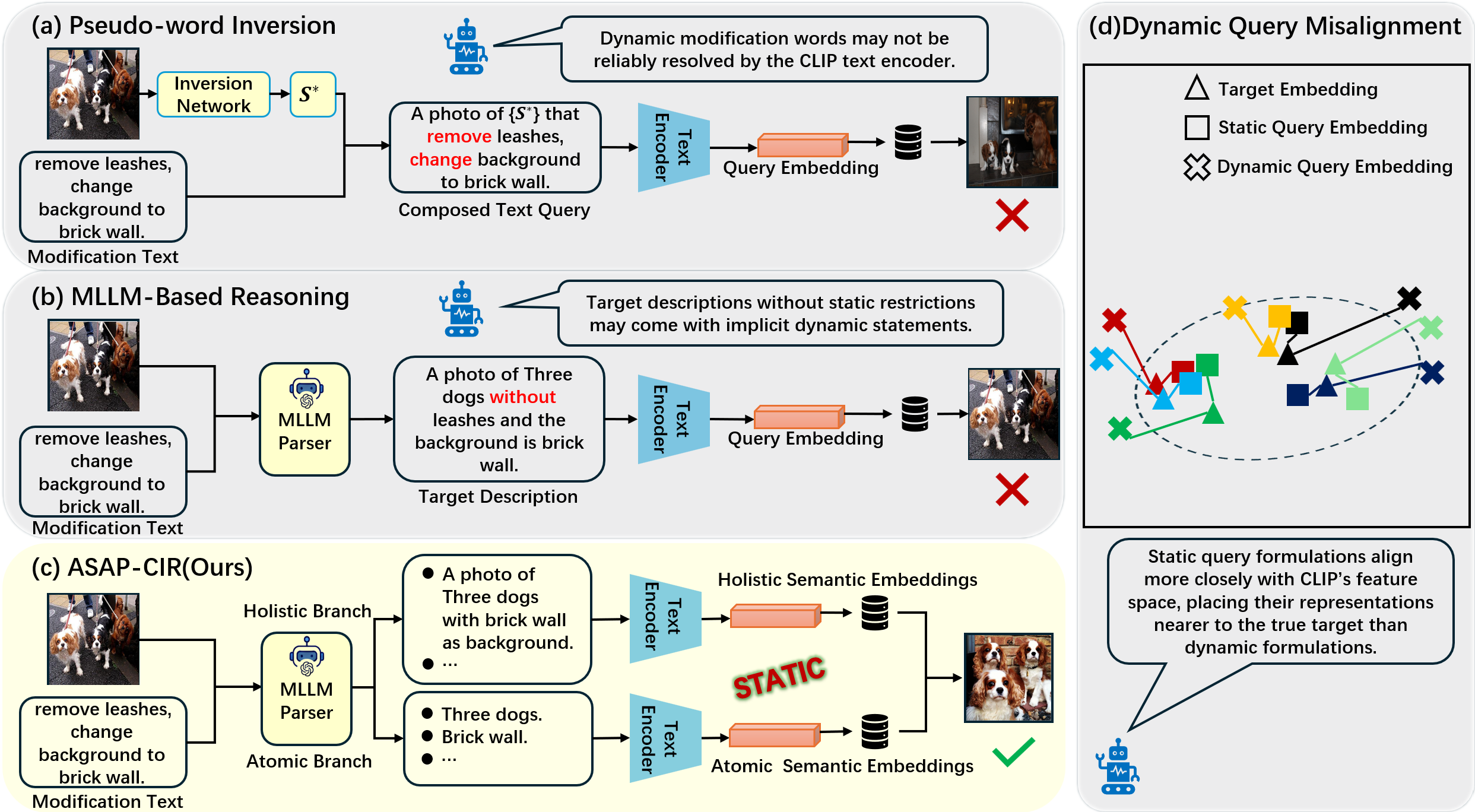}
    \caption{
    Motivation of ASAP-CIR. The left panels compare three query formulation paradigms: (a) pseudo-word inversion directly combines an inverted visual token with dynamic modification text; (b) MLLM-based reasoning generates a target description without explicitly constraining it to static visual semantics; and (c) ASAP-CIR reconstructs the composed query as complementary holistic descriptions and importance-weighted atomic semantics for multi-level retrieval. Panel (d) visualizes the phenomenon observed in our diagnostic experiments: target-side static query embeddings are more closely aligned with the target embedding region, whereas dynamic composed query embeddings are more dispersed and frequently deviate from the intended target state.
    }
    \label{fig:motivation}
\end{figure}

We instantiate this formulation with ASAP-CIR. ASAP abbreviates \textbf{A}tomic \textbf{S}tatic sem\textbf{A}ntic \textbf{P}arsing, and ASAP-CIR is a training-free framework for ZS-CIR. As illustrated on the left of Fig.~\ref{fig:motivation}, a frozen MLLM reconstructs each dynamic query as a static target representation composed of multiple holistic descriptions and importance-weighted atomic semantics. The holistic descriptions capture target identity and scene-level coherence; the atomic semantics expose objects, attributes, relations, counts, styles, and scene cues as explicit retrieval constraints. Candidate images are ranked using holistic state alignment and atomic constraint grounding, followed by calibrated target-state evidence aggregation. The right panel summarizes the diagnostic evidence motivating this design: target-side static query representations tend to occupy the target-aligned region of CLIP's feature space more consistently than dynamic composed query representations.

To examine whether this reformulation is supported by the behavior of pretrained vision-language representations, we further conduct a controlled text-only diagnostic study. Under this protocol, static target descriptions retrieve the intended target descriptions more reliably than formulations that combine a reference description and modification instruction. This result does not imply a universal inability of CLIP to encode dynamic language; instead, it shows that query formulation materially affects retrieval alignment in the considered CIR setting. Benchmark experiments further reveal that the benefit is dataset dependent: it is most pronounced on CIRCO, while FashionIQ and CIRR expose limitations in fine-grained garment discrimination and deeper candidate ranking.

The main contributions of this work are summarized as follows:
\begin{itemize}
    \item We identify and formulate a representation mismatch in ZS-CIR: transformation-oriented instructions are used to query pretrained image-text spaces whose candidates express completed visual states. We consequently reformulate ZS-CIR as target-state reconstruction followed by retrieval.
    \item We introduce a multi-level static target representation that jointly preserves holistic semantics and fine-grained atomic constraints. ASAP-CIR provides a training-free realization of this principle and a structured interface between frozen MLLM reasoning and pretrained VLM retrieval.
    \item We establish controlled evidence that target-state formulations are better aligned with pretrained CLIP representations for several transformation types, and evaluate the resulting retrieval behavior on FashionIQ, CIRR, and CIRCO, including its benefits and remaining boundaries.
\end{itemize}

\section{Related Work}
\label{sec:related_work}

\subsection{Composed Image Retrieval}
\label{subsec:related_cir}
Composed image retrieval retrieves images according to a reference image and a natural-language modification. Unlike conventional image retrieval, in which a query is expressed through a single modality, CIR requires the retrieval model to preserve query-relevant visual content from the reference while satisfying the changes expressed in language. Existing CIR methods can be broadly divided into supervised and zero-shot paradigms. Supervised methods learn this composition from annotated triplets of reference images, modification texts, and target images. Early approaches such as TIRG \citep{Vo_2019_CVPR} learn dedicated feature composition modules to combine visual and textual cues, while later methods further exploit pretrained vision-language models, cross-modal attention, or task-specific pretraining objectives to improve composed retrieval performance \citep{Baldrati_2022_CVPR,Han_2023_CVPR,Song_2024_CVPR,Fu_2026_CVPR}. Although these methods achieve strong performance when training triplets are available, their reliance on annotated CIR data limits their scalability to new domains and retrieval scenarios. In addition, supervised models tend to inherit the semantic distribution of the training triplets, making it difficult to handle rare or open-ended modification intents without collecting new annotations.

Several benchmarks have been proposed to support CIR research. Fashion200k \citep{Han_2017_ICCV} is an early fashion-domain benchmark that connects product images with relative textual feedback. FashionIQ \citep{Wu_2021_CVPR} further provides human-written natural language feedback for fashion retrieval across multiple garment categories. CIRR \citep{Liu_2021_ICCV} extends composed retrieval to real-life open-domain images and introduces both full-gallery and subset evaluation protocols, while CIRCO \citep{Baldrati_2023_ICCV} builds a challenging COCO-based benchmark with multiple valid target images for each query.

To reduce the dependence on CIR triplets, recent studies have increasingly focused on ZS-CIR. One representative line of work follows the textual inversion paradigm. Pic2Word \citep{Saito_2023_CVPR} maps an input image into a pseudo-word token in the CLIP text embedding space, enabling the composed query to be represented as a text prompt. SEARLE \citep{Baldrati_2023_ICCV} improves this idea through a textual inversion mechanism that represents the reference image as a pseudo-word token and composes it with the relative caption. Subsequent methods such as LinCIR \citep{Gu_2024_CVPR} reduce the need for image-text supervision by using language-only training, while DeG \citep{CHEN2026113187}, MOA \citep{10.1145/3726302.3730074}, and CoCo-IR \citep{cao2026cocoir} further study data-efficient generalization, object-aware pseudo-word learning, and contextual composed retrieval. Beyond textual inversion, recent ZS-CIR methods also explore difference-aware composition, semantic flow, geodesic semantic expansion, and reward-guided test-time optimization, as represented by DiffComp, FlowCIR, G-MIXER, and TT-RLDR \citep{Zhang_2026_CVPR,he2026flowcir,lim2026gmixer,Zhou2026TTRLDR}. These methods are elegant and efficient because they reuse or adapt the pretrained CLIP image-text alignment space. In textual inversion methods, however, the pseudo-word representation of the reference image is typically inserted into a prompt together with the original modification text. Consequently, explicit transition expressions such as removal, replacement, or attribute change may remain in the composed prompt and be passed directly to the CLIP text encoder. Such methods therefore implicitly assume that CLIP can interpret these dynamic modification semantics and combine them correctly with the pseudo-word representation.

Another recent line of work explores training-free ZS-CIR with LLMs or MLLMs. CIReVL \citep{karthik2024visionbylanguage} first captions the reference image with a VLM and then uses an LLM to rewrite the composed query into target descriptions for CLIP retrieval. LDRE \citep{10.1145/3626772.3657740} further generates dense reference captions and performs LLM-based divergent reasoning, followed by semantic-relevance-weighted caption ensemble. PrediCIR \citep{Tang_2025_PrediCIR} predicts missing target-relevant information with a world model, and CoTMR \citep{Sun_2025_ICCV} introduces chain-of-thought multi-scale reasoning for training-free retrieval. AutoCIR \citep{10.1145/3711896.3736982} formulates ZS-CIR as automatic multi-agent collaboration with planning, retrieval, and correction agents, while OSrCIR \citep{Tang_2025_OSrCIR} employs one-stage reflective chain-of-thought reasoning with an MLLM before retrieval. More recent methods such as STiTch \citep{Li_2026_CVPR} and WISER \citep{wang2026wiser} further improve training-free retrieval through semantic transition and transportation, wider search, deeper reasoning, or adaptive fusion. These approaches demonstrate that external reasoning can reduce reliance on directly encoding raw modification text. Because some of them infer descriptions of the target, their outputs may incidentally avoid explicit transition words and contain implicit static target semantics. However, static semantic formulation is generally a by-product rather than an explicit constraint: generated descriptions are not necessarily required to exclude source-state or transition-oriented expressions, nor are target requirements systematically organized at complementary semantic levels. Their primary emphasis is instead on composition, reasoning, search, or fusion.

Accordingly, the distinction of ASAP-CIR is not merely the use of an MLLM or an additional patch-level matching branch. Existing methods mainly investigate how reference and modification information should be composed, reasoned over, or refined; our work instead asks how the inferred target state should be represented before retrieval. ASAP-CIR answers this question through a multi-level static target representation that combines holistic descriptions with importance-weighted atomic semantics, thereby making query representation itself an explicit component of retrieval design.

\subsection{Vision-Language Models for CIR}
\label{subsec:related_vlm}
Vision-language models (VLMs) have become a fundamental component of modern multimodal learning, as they provide a shared representation space for visual and textual information. Among them, CLIP, introduced by Radford et al. \citep{pmlr-v139-radford21a}, is one of the most representative models for open-vocabulary visual understanding. CLIP is trained on large-scale image-text pairs using a contrastive learning objective, where matched image-text pairs are pulled closer and mismatched pairs are pushed apart in a shared embedding space. This simple yet scalable pretraining strategy enables CLIP to learn transferable visual representations and perform zero-shot recognition by comparing image features with natural language prompts.

Due to its strong image-text alignment ability, CLIP has been widely adopted as a general-purpose backbone for many downstream tasks, including zero-shot and few-shot image classification, test-time adaptation, robust recognition, anomaly detection, image-language understanding, visual question answering, and compositional zero-shot learning \citep{Martin_2024_CVPR,Saha_2024_CVPR,Zanella_2024_CVPR,Ge_2023_CVPR,ICLR2024_8e5b9dc3,ICLR2024_7c40c505,ICLR2025_2c2be726,ICLR2026_cb825b4e,Jeong_2023_CVPR,Tschannen_2023_CVPR,Guo_2023_CVPR,Chen_2024_CVPR,Lu_2023_CVPR,Huang_2024_CVPR,Wu_2025_CVPR}. These studies demonstrate the broad transferability of pretrained vision-language representations. At the same time, they also suggest that the way a task is formulated in language can strongly affect how CLIP is used. This issue is particularly relevant to CIR, where the query text often specifies how one visual state should be transformed into another rather than simply describing a static image.

In composed image retrieval, CLIP is particularly attractive because the task naturally requires matching visual content with language descriptions. Existing ZS-CIR methods often reuse CLIP as the core retrieval backbone, either by mapping the reference image into a pseudo-word token, generating target captions, or composing reference and modification semantics in the CLIP feature space \citep{Saito_2023_CVPR,Baldrati_2023_ICCV,karthik2024visionbylanguage,10.1145/3626772.3657740}. Their success nevertheless depends on the ability of the pretrained text encoder to represent transformation-oriented language in a form suitable for matching completed visual states. This issue is relevant to negation, counting, relational changes, and object removal, where a query may mention concepts or attributes that should not appear in the target. In this work, we retain CLIP as the retrieval backbone but study whether reformulating the query as observable target-state semantics provides a more suitable retrieval representation under a controlled CIR protocol.

\section{Method}
\label{sec:method}

This section presents ASAP-CIR as a concrete training-free solution to the query representation problem in zero-shot composed image retrieval (ZS-CIR). We use dynamic semantics to denote instructions that specify a transition from the reference state, and target-state semantics to denote visually observable content in the desired result. Motivated by the controlled evidence in Sec.~\ref{subsec:clip_dynamic_analysis}, ASAP-CIR follows a representation-centered pipeline: it reconstructs the intended target state from the composed query, expresses that state through complementary holistic descriptions and importance-weighted atomic semantics, and grounds both levels against candidate images. CLIP is therefore used primarily to align observable target-state semantics rather than to infer the transformation itself.

\begin{figure*}[t]
    \centering
    \includegraphics[width=0.96\textwidth]{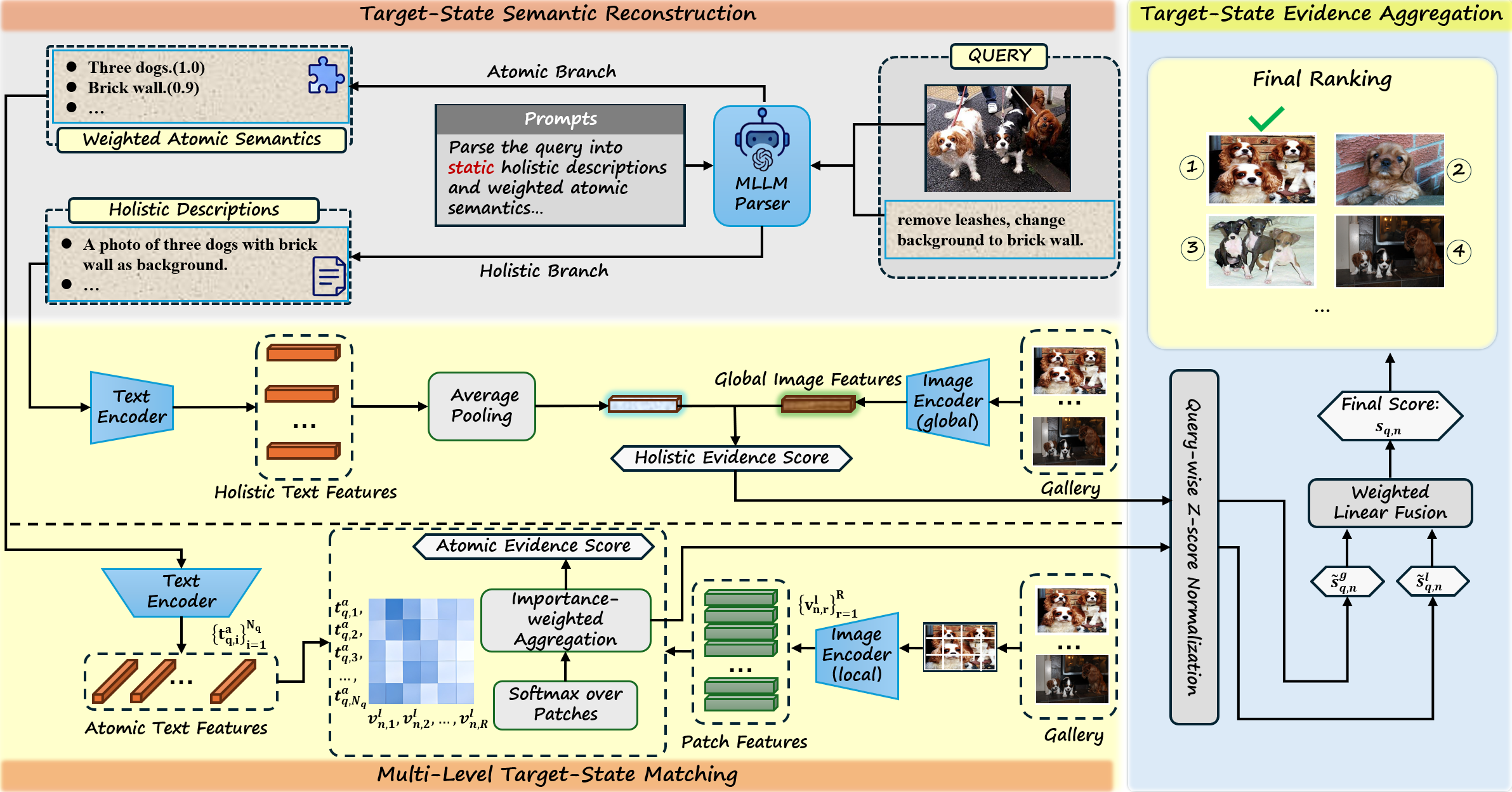}
    \caption{
    Overview of the proposed ASAP-CIR framework. Given a composed query, the MLLM parser performs target-state semantic reconstruction and produces complementary holistic descriptions and importance-weighted atomic semantics. In the figure, these outputs are labeled ``Holistic Descriptions'' and ``Weighted Atomic Semantics,'' respectively. The holistic descriptions are encoded independently and combined by average pooling before being aligned with global image features to obtain the holistic evidence score. The atomic semantics are encoded independently and grounded against candidate patch features through patch-wise softmax assignment and importance-weighted aggregation, producing the atomic evidence score. Finally, query-wise z-score normalization calibrates both evidence scores, and weighted linear fusion produces the final retrieval score and ranking.
    }
    \label{fig:method_overview}
\end{figure*}

We first formulate the ZS-CIR task in Sec.~\ref{subsec:problem_definition}. As illustrated in Fig.~\ref{fig:method_overview}, ASAP-CIR proceeds through three computational stages: target-state semantic reconstruction, multi-level target-state matching, and target-state evidence aggregation. For methodological organization, matching and evidence aggregation are presented together in Sec.~\ref{subsec:global_local_matching} because both operate on the reconstructed representation without introducing additional trainable components.

The first module, \textbf{Target-State Semantic Reconstruction}, uses an off-the-shelf MLLM to infer the completed visual state and express it at holistic and atomic levels. Holistic descriptions summarize the expected appearance and scene, whereas atomic semantics preserve fine-grained objects, attributes, relations, and other observable requirements. This stage moves transformation reasoning outside the CLIP text encoder and is described in Sec.~\ref{subsec:dynamic_to_static}.

The second module, \textbf{Multi-Level Target-State Matching}, evaluates each candidate using complementary evidence. Holistic state alignment measures overall image--text consistency, while atomic constraint grounding measures how strongly candidate patches support individual target-state facts. Their normalized scores are then aggregated to produce the final ranking. We present this process in Sec.~\ref{subsec:global_local_matching}.

\subsection{Problem Definition}
\label{subsec:problem_definition}

We consider the training-free ZS-CIR setting, in which no annotated triplets of reference images, modification texts, and target images are used for optimization. All pretrained model parameters remain frozen, and retrieval is performed without task-specific training.

Given a reference image $I_r$ and a modification text $T_m$, the goal of composed image retrieval (CIR) is to rank images from a candidate gallery $\mathcal{G}$ according to their consistency with the intended target semantics. The reference image provides the original visual content, while the modification text specifies the desired semantic changes relative to the reference image. Together, they form a composed query $q=(I_r,T_m)$. Given a candidate gallery $\mathcal{G}=\{I_1,I_2,\ldots,I_N\}$, CIR aims to assign higher matching scores to valid target images than to irrelevant candidates. This formulation covers both single-target benchmarks such as FashionIQ and CIRR and multi-target benchmarks such as CIRCO. With $S(\cdot,\cdot)$ denoting the matching function between the composed query and a candidate image, retrieval can be viewed as ranking all candidates according to $S(q,I_i)$ and returning the top-ranked images. A desirable retrieval model should therefore assign high scores to candidate images that both preserve the query-relevant visual content of the reference image and satisfy the semantic changes specified by the modification text.

\subsection{Target-State Semantic Reconstruction}
\label{subsec:dynamic_to_static}

Given a composed query $q=(I_r,T_m)$, where $I_r$ denotes the reference image and $T_m$ denotes the modification text, the first module uses a frozen MLLM to jointly interpret both inputs and reconstruct the visual content expected in the target image. It expresses this inferred state through complementary holistic descriptions and atomic semantics without updating any model parameters. Formally, the reconstruction process is written as

\begin{equation}
(\mathcal{D}_q,\mathcal{A}_q)=
\Phi_{\mathrm{MLLM}}
\left(
I_r,T_m;\mathcal{P}_d
\right),
\label{eq:mllm_static_parsing}
\end{equation}

\noindent
where $\Phi_{\mathrm{MLLM}}$ denotes the frozen MLLM and $\mathcal{P}_d$ denotes the dataset-specific prompt used to express the inferred target state in a visually observable form. The output contains a holistic description set $\mathcal{D}_q$ and an importance-weighted atomic semantic set $\mathcal{A}_q$. The former captures the overall appearance and scene-level state, while the latter decomposes that state into fine-grained visual requirements.

\subsubsection{Static Semantic Canonicalization}

The prompt $\mathcal{P}_d$ is designed to enforce two constraints. First, the generated text should describe the final target image directly, rather than explaining the editing process from the reference image. Second, each atomic semantic should correspond to a visually observable static fact, such as an object, attribute, relation, material, count, style, or scene cue. In other words, expressions such as ``remove the dog'' or ``change the bottle to orange'' are discouraged, while static formulations such as ``a single dog is visible'' or ``an orange soft drink bottle is present'' are preferred. This explicit constraint is important because it aligns the generated text with the static image-caption form on which CLIP-like models are pretrained.

We use dataset-specific prompts because the preservation assumptions are different across benchmarks. In FashionIQ, the reference garment usually provides important identity and style information, so unchanged garment attributes should be retained unless they conflict with the modification. In CIRR and CIRCO, the modification may explicitly add, remove, or replace objects and scene context, so the MLLM is asked to infer which reference details should remain visible and which should be suppressed. This prompt design allows the parser to produce target semantics that are both static and adapted to each query.

\subsubsection{Holistic Target-State Representation}
To obtain a robust holistic representation of the target image, the MLLM generates $K$ semantically consistent yet linguistically diverse holistic descriptions:

\begin{equation}
\mathcal{D}_q
=
\left\{
d_q^{k}
\right\}_{k=1}^{K},
\label{eq:global_descriptions}
\end{equation}

\noindent
where $d_q^{k}$ denotes the $k$-th holistic description and $K$ is the number of generated descriptions.

Each $d_q^{k}$ is formulated as a standalone static image caption that directly describes the expected target image, rather than as an editing instruction describing how the reference image should be modified. In practice, we prompt the model to generate caption-style expressions, such as sentences beginning with ``A photo of'' or ``A product photo of''. These descriptions characterize the target image from a global perspective, including its primary objects, attributes, scene context, and overall visual appearance.

Generating multiple holistic descriptions provides complementary linguistic views of the same target semantics. Although these descriptions are semantically consistent, different expressions may emphasize different aspects of the target image. This reduces the risk that a single generated caption omits important discriminative information and provides a more comprehensive holistic representation for subsequent retrieval. More importantly, all generated descriptions remain in the form of static image captions, making them naturally compatible with the image--text matching paradigm of CLIP.

\subsubsection{Atomic Target-State Representation}
Although the holistic descriptions provide an overall representation of the expected target image, they may still fail to explicitly preserve all fine-grained requirements contained in the composed query. This limitation becomes more evident when the modification text involves multiple objects, attributes, spatial relations, stylistic changes, or other local visual constraints. Therefore, in parallel with holistic description generation, we further require the MLLM to decompose the inferred target semantics into a variable-size set of importance-weighted atomic semantics:

\begin{equation}
\mathcal{A}_q = \{(a_q^{i}, \alpha_q^{i})\}_{i=1}^{N_q}.
\label{eq:static_semantic_units}
\end{equation}

\noindent
where $a_q^{i}$ denotes the $i$-th atomic semantic, $\alpha_q^{i}\in[0,1]$ denotes its corresponding importance weight, and $N_q$ is the number of atomic semantics generated for query $q$. Each atomic semantic is expressed as an independent static visual description that captures a specific fine-grained requirement of the desired target image. These requirements may correspond to objects, attributes, spatial relations, styles, counts, materials, scenes, or other visually observable constraints implied by the composed query.

The importance weight $\alpha_q^{i}$ is a heuristic score estimated by the MLLM that indicates the relative retrieval relevance of the $i$-th semantic requirement. Semantics that are directly related to the requested modification or are highly discriminative for the target are assigned larger weights, whereas less informative contextual details receive smaller weights. We do not treat these values as calibrated probabilities; instead, they provide a lightweight query prior for emphasizing more important visual requirements during atomic constraint grounding.

Since $N_q$ is query-dependent, the proposed representation naturally adapts to different levels of semantic complexity. A simple query may require only a small number of atomic semantics, whereas a more complex modification text can be decomposed into a larger set of static visual requirements. This variable-size representation avoids forcing heterogeneous queries into a fixed semantic structure and allows ASAP-CIR to preserve query-specific information more faithfully.

The importance-weighted atomic semantic set $\mathcal{A}_q$ is directly used in the subsequent atomic constraint grounding stage. Unlike the holistic descriptions in $\mathcal{D}_q$, which characterize the expected target image from an overall perspective, $\mathcal{A}_q$ focuses on explicit fine-grained constraints that should be satisfied by the retrieved image. The term ``atomic'' indicates that each unit describes a compact static visual fact. The two representations therefore provide complementary views of the target semantics: $\mathcal{D}_q$ captures holistic semantic consistency, while $\mathcal{A}_q$ preserves fine-grained atomic requirements.

Overall, the proposed parsing module transforms the original composed query into two complementary static target representations: the holistic description set $\mathcal{D}_q$, which summarizes the overall semantics of the expected target image, and the importance-weighted atomic semantic set $\mathcal{A}_q$, which explicitly represents its fine-grained visual constraints. These two representations are subsequently used for holistic state alignment and atomic constraint grounding, respectively.

The MLLM operates only on the query side; candidate images are represented directly by CLIP. Their features can therefore be precomputed offline, avoiding gallery-side MLLM inference while preserving training-free retrieval efficiency.

\subsection{Multi-Level Target-State Matching}
\label{subsec:global_local_matching}
After reconstructing the target-state representation, ASAP-CIR ranks images using complementary holistic and atomic evidence. Let $\mathcal{G}=\{I_n\}_{n=1}^{N}$ denote the candidate gallery. We use a frozen vision-language model with an image encoder $E_v$ and a text encoder $E_t$ to compute candidate-level holistic similarities and atomic-to-patch similarities, from which complementary holistic and atomic evidence scores are derived.

\subsubsection{Holistic State Alignment}
For each query $q$, the holistic state alignment branch uses the generated holistic description set $\mathcal{D}_q=\{d_q^k\}_{k=1}^{K}$, where $K$ denotes the number of holistic descriptions used to represent the overall target semantics. These descriptions provide multiple static textual views of the desired target image and are jointly encoded to form a robust holistic query representation.

Each holistic description is encoded by the text encoder. We first normalize each text feature, average the normalized features, and then normalize the averaged feature to obtain the holistic query representation:
\begin{equation}
\mathbf{t}_q^{g}
=
\mathrm{Norm}
\left(
\frac{1}{K}
\sum_{k=1}^{K}
\mathrm{Norm}
\left(
E_t(d_q^k)
\right)
\right),
\label{eq:global_query_feature}
\end{equation}
\noindent
where $\mathrm{Norm}(\cdot)$ denotes $\ell_2$ normalization. For each candidate image $I_n$, we extract its global image feature as
\begin{equation}
\mathbf{v}_n^{g}
=
\mathrm{Norm}
\left(
E_v^{g}(I_n)
\right),
\label{eq:global_image_feature}
\end{equation}
\noindent
where $E_v^{g}(\cdot)$ denotes the global image representation produced by the image encoder. The holistic evidence score is computed by cosine similarity:
\begin{equation}
s_{q,n}^{g}
=
\mathbf{t}_q^{g}
(\mathbf{v}_n^{g})^{\top}.
\label{eq:global_score}
\end{equation}

\subsubsection{Atomic Constraint Grounding}
For atomic constraint grounding, each query is represented by the importance-weighted atomic semantic set $\mathcal{A}_q=\{(a_q^i,\alpha_q^i)\}_{i=1}^{N_q}$. The number of atomic semantics is query dependent. Before grounding, malformed and empty entries are discarded, while all remaining valid atomic entries are retained without forced truncation. Consequently, $N_q$ denotes the variable number of valid atomic semantics retained for query $q$. Each $a_q^i$ is a static visual phrase or short caption, and $\alpha_q^i\in[0,1]$ is its importance score.

Each atomic semantic is encoded by the text encoder:
\begin{equation}
\mathbf{t}_{q,i}^{a}
=
\mathrm{Norm}
\left(
E_t(a_q^i)
\right).
\label{eq:atomic_text_feature}
\end{equation}
\noindent
For each candidate image $I_n$, we extract a set of patch features from the image encoder:
\begin{equation}
\mathcal{V}_n^{l}
=
\{\mathbf{v}_{n,r}^{l}\}_{r=1}^{R},
\quad
\mathbf{v}_{n,r}^{l}
=
\mathrm{Norm}
\left(
E_v^{l}(I_n)_r
\right),
\label{eq:local_image_feature}
\end{equation}
\noindent
where $R$ is the number of visual patch tokens and $E_v^{l}(\cdot)$ denotes the patch-token feature output of the image encoder used for atomic constraint grounding.

We compute the similarity between each atomic semantic and each candidate patch feature:
\begin{equation}
m_{q,n}^{i,r}
=
\mathbf{t}_{q,i}^{a}
(\mathbf{v}_{n,r}^{l})^{\top}.
\label{eq:atom_patch_similarity}
\end{equation}
\noindent
Then, a softmax operation is applied over all patch features of the candidate image:
\begin{equation}
\rho_{q,n}^{i,r}
=
\frac{
\exp(m_{q,n}^{i,r}/\tau)
}{
\sum_{r'=1}^{R}
\exp(m_{q,n}^{i,r'}/\tau)
},
\label{eq:softmax_patch_weight}
\end{equation}
\noindent
where $\tau$ is the atomic grounding temperature. The soft semantic coverage score of the $i$-th atomic semantic with respect to candidate image $I_n$ is computed as
\begin{equation}
c_{q,n}^{i}
=
\sum_{r=1}^{R}
\rho_{q,n}^{i,r}
m_{q,n}^{i,r}.
\label{eq:atomic_coverage_score}
\end{equation}

\noindent
This score provides soft evidence that the candidate supports the corresponding atomic target-state constraint through its most relevant patches. It is not treated as binary logical satisfaction, since both the text--patch similarities and their assignment remain continuous.
Before aggregating atomic semantics, we normalize their importance scores:
\begin{equation}
\hat{\alpha}_q^i
=
\frac{
\max(\alpha_q^i,0)
}{
\sum_{j=1}^{N_q}
\max(\alpha_q^j,0)
+\epsilon_{\alpha}
},
\label{eq:importance_normalization}
\end{equation}
\noindent
where $\epsilon_{\alpha}>0$ is a small constant used for numerical stability. The atomic evidence score is then obtained by importance-weighted aggregation:
\begin{equation}
s_{q,n}^{l}
=
\sum_{i=1}^{N_q}
\hat{\alpha}_q^i
c_{q,n}^{i}.
\label{eq:local_score}
\end{equation}

\subsubsection{Target-State Evidence Aggregation}
Finally, evidence aggregation combines the holistic and atomic scores linearly to obtain the final retrieval score. In our implementation, query-wise z-score normalization is enabled by default. Therefore, before aggregation, both scores are normalized over the candidate gallery for each query:
\begin{equation}
\tilde{s}_{q,n}^{g}
=
\frac{
s_{q,n}^{g}
-
\mu_q^{g}
}{
\max(\sigma_q^{g},\epsilon_z)
},
\label{eq:global_score_normalization}
\end{equation}
\begin{equation}
\tilde{s}_{q,n}^{l}
=
\frac{
s_{q,n}^{l}
-
\mu_q^{l}
}{
\max(\sigma_q^{l},\epsilon_z)
},
\label{eq:local_score_normalization}
\end{equation}
\noindent
where $\mu_q^{g}$ and $\sigma_q^{g}$ denote the mean and standard deviation of the holistic scores for query $q$, $\mu_q^{l}$ and $\sigma_q^{l}$ denote the corresponding statistics of the atomic scores, and $\epsilon_z>0$ is a small constant that prevents division by zero. The final retrieval score is then computed as
\begin{equation}
s_{q,n}
=
\lambda_g \tilde{s}_{q,n}^{g}
+
\lambda_l \tilde{s}_{q,n}^{l},
\label{eq:final_score}
\end{equation}
\noindent
where $\lambda_g$ and $\lambda_l$ weight the holistic and atomic evidence, respectively. Candidate images are ranked according to $s_{q,n}$, and the top-ranked images are returned as retrieval results.

\section{Experiments}
\label{sec:experiments}

\subsection{Experimental Setup}
\label{subsec:datasets_metrics}

\textbf{Datasets.}
Following recent ZS-CIR studies \citep{lim2026gmixer,Li_2026_CVPR}, we evaluate ASAP-CIR on three widely used composed image retrieval benchmarks: FashionIQ \citep{Wu_2021_CVPR}, CIRR \citep{Liu_2021_ICCV}, and CIRCO \citep{Baldrati_2023_ICCV}. FashionIQ is a fashion-domain retrieval benchmark containing three categories, namely Dress, Shirt, and Toptee. We evaluate on the validation split and report recall at rank $k$ (R@$k$), with $k\in\{10,50\}$, for each category, together with their average over the three categories. CIRR is an open-domain benchmark constructed from real-life images sourced from NLVR2. It augments reference--target image pairs with human-written relative captions that describe the intended semantic change, thereby emphasizing fine-grained discrimination among visually similar natural images. CIRR provides both full-gallery and visually restricted subset evaluation protocols. We evaluate on the test1 split and report full-gallery R@$k$ with $k\in\{1,5,10,50\}$; to keep the main comparison focused on this setting, we do not report subset recall. CIRCO is built from the COCO 2017 unlabeled image set and provides multiple ground-truth target images for each query. It contains 123,403 candidate images and 800 test queries. Following the official evaluation protocol, we report mean average precision at rank $k$ (mAP@$k$), with $k\in\{5,10,25,50\}$.

\textbf{Baselines.}
We compare ASAP-CIR with representative zero-shot composed image retrieval methods using reported backbone architecture groups whenever available. The compared methods include training-based or language-supervised approaches, such as Pic2Word \citep{Saito_2023_CVPR}, SEARLE \citep{Baldrati_2023_ICCV}, LinCIR \citep{Gu_2024_CVPR}, FlowCIR \citep{he2026flowcir}, DiffComp \citep{Zhang_2026_CVPR}, and TT-RLDR \citep{Zhou2026TTRLDR}. We also compare with recent training-free methods that rely on foundation models or test-time reasoning, including CIReVL \citep{karthik2024visionbylanguage}, LDRE \citep{10.1145/3626772.3657740}, AutoCIR \citep{10.1145/3711896.3736982}, OSrCIR \citep{Tang_2025_OSrCIR}, and STiTch \citep{Li_2026_CVPR}. In the tables, T-Free denotes whether a method is training-free. We group methods by their reported visual encoder architecture for readability, while noting that exact pretrained checkpoints, preprocessing pipelines, and implementation details may differ across papers. Following common practice, the best and second-best results within each backbone group are shown in \textbf{bold} and \underline{underlined}, respectively.

\textbf{Implementation Details.}
ASAP-CIR is a training-free method and does not use any CIR triplets for optimization. We use Qwen3.6-27B as the frozen MLLM for static semantic parsing of queries. In our implementation, the number of holistic descriptions is controlled by the hyperparameter \texttt{num-target-descriptions}, and we set $K=3$ by default. If fewer than $K$ valid descriptions are produced, the last valid description is repeated until the number reaches $K$; if more than $K$ descriptions are produced, only the first $K$ descriptions are retained. The prompt requests 4--7 semantic lines, but generation is not constrained to an exact count. All valid atomic lines are retained without forced truncation, so the resulting $N_q$ remains query dependent. Candidate images are not processed by the MLLM; instead, their visual features are extracted and cached before retrieval. For CLIP-based matching, we use OpenCLIP backbones, including ViT-B/32, ViT-L/14, and ViT-G/14, with LAION pretrained weights. Unless otherwise specified, ViT-L/14 with \texttt{laion2b\_s32b\_b82k} is used as the default backbone. We apply \texttt{TargetPad} preprocessing with a \texttt{targetpad-ratio} of 1.25 and perform feature extraction in \texttt{float16} precision.

During MLLM inference, all generated results are saved as structured query semantic files and reused for retrieval. This avoids repeated MLLM calls when testing different fusion weights, temperatures, or CLIP backbones. For holistic state alignment, the three generated holistic descriptions are encoded independently and then averaged after feature normalization. For atomic constraint grounding, each atomic semantic is encoded as a short static text query and matched against patch features extracted from the candidate image encoder. All candidate image features are cached under the project directory, so changing retrieval hyperparameters does not require re-encoding the image gallery.

\textbf{Hardware and retrieval settings.}
MLLM parsing is performed on six NVIDIA GeForce RTX 3090 GPUs, while retrieval and feature scoring are performed on a single RTX 3090 GPU. Candidate image features are cached in advance, so retrieval only requires text encoding, candidate-feature loading, and matrix-based similarity computation. We use an atomic grounding temperature of 0.07. The holistic and atomic evidence scores undergo query-wise z-score normalization before fusion. Unless otherwise stated, the holistic/atomic fusion weights $(\lambda_g,\lambda_l)$ are set to 0.90/0.10 for FashionIQ, 0.85/0.15 for CIRR, and 0.95/0.05 for CIRCO. These dataset-specific weights are selected from coarse preliminary sweeps and then fixed for the reported runs; the CIRCO sensitivity analysis in Sec.~\ref{subsec:ablation_study} makes this choice transparent.

\subsection{Benchmark Performance}
\label{subsec:benchmark_comparisons}

We compare ASAP-CIR with representative ZS-CIR methods on FashionIQ, CIRR, and CIRCO. The results are reported in Table~\ref{tab:fashioniq_results}, Table~\ref{tab:cirr_results}, and Table~\ref{tab:circo_results}, respectively. Overall, ASAP-CIR achieves competitive performance across the three benchmarks and shows its clearest advantage on CIRCO, where multiple ground-truth targets require a retrieval model to capture diverse yet accurate target semantics.

\subsubsection{FashionIQ}
Table~\ref{tab:fashioniq_results} presents the results on FashionIQ. With ViT-B/32, ASAP-CIR obtains 30.99 average R@10 and 53.20 average R@50; with ViT-L/14, it reaches 35.00 and 55.02, respectively. Within the reported comparison set, the method performs particularly well on Dress and Toptee under ViT-L/14, suggesting that the parsed atomic semantics can capture discriminative attributes such as color, texture, sleeve length, silhouette, and style. Its lower Shirt scores indicate that static semantic parsing does not uniformly resolve all fine-grained fashion retrieval cases, particularly when visually similar garments differ in subtle structural details.

It is worth noting that FashionIQ descriptions are usually short and attribute-centric. A single holistic description may already capture the dominant garment category, but it may underrepresent subtle style changes that determine the correct target among visually similar products. Atomic evidence complements holistic evidence by assigning explicit weights to these fine-grained attributes. Therefore, even without any training on FashionIQ triplets, ASAP-CIR can remain competitive with methods that rely on learned textual inversion or task-specific composition.

\subsubsection{CIRR}
Table~\ref{tab:cirr_results} reports the results on CIRR. ASAP-CIR obtains 28.41 R@1 with ViT-B/32 and 32.80 R@1 with ViT-L/14. The latter exceeds AutoCIR by 0.99 points under the same backbone grouping. However, ASAP-CIR becomes less competitive at larger retrieval depths, where DiffComp and AutoCIR obtain stronger results. This behavior may reflect the emphasis of our scoring function on precise semantic satisfaction rather than explicit candidate-pool expansion. CIRR therefore remains a challenging setting for the current formulation, particularly when diverse candidates share similar global content.

This behavior may be related to the design of our method. Since ASAP-CIR converts the query into explicit target semantic requirements, candidates that violate key atomic constraints tend to be penalized during atomic constraint grounding. Such scoring can help select a highly precise top-ranked image, but may reduce the diversity of lower-ranked candidates. We consider this a reasonable trade-off for CIRR, where the top-1 result is particularly important for evaluating whether the composed query is correctly grounded.

\subsubsection{CIRCO}
As shown in Table~\ref{tab:circo_results}, ASAP-CIR obtains strong results on CIRCO under both ViT-L/14 and ViT-G/14 backbones. With ViT-L/14, it reaches 35.49, 36.46, 39.21, and 40.35 mAP at $k=\{5,10,25,50\}$, improving over STiTch by 9.94, 10.19, 10.40, and 10.36 points, respectively. The improvement is consistent across all retrieval depths. With the stronger ViT-G/14 backbone, ASAP-CIR reaches 37.06, 38.47, 41.44, and 42.52 mAP, exceeding STiTch by 2.66, 2.91, 3.37, and 2.50 points. These results indicate that converting dynamic modification requirements into explicit holistic descriptions and atomic semantics provides an effective retrieval signal for CIRCO, where the target images may satisfy the query in multiple plausible ways.

The consistent advantage on CIRCO is aligned with our motivation. Unlike single-target evaluation, mAP rewards methods that can rank multiple semantically valid images ahead of distractors. The holistic descriptions generated by the MLLM provide broad target coverage, while the importance-weighted atomic semantics help suppress candidates that match only the reference image or only a subset of the required changes. This combination is particularly beneficial for queries involving multiple acceptable targets, since the retrieval model must capture the common static semantics shared by all valid answers rather than overfitting to one textual expression.

\input{experiment_tables}

\subsection{Target-State Alignment Analysis in Pretrained VLMs}
\label{subsec:clip_dynamic_analysis}

Beyond benchmark comparisons, we test the central representation hypothesis directly: does a description of the completed target state align with target candidates more reliably than a transformation-oriented composed query? This diagnostic does not claim a universal inability of CLIP to understand dynamic language. Instead, it isolates a controlled textual setting relevant to CIR and measures the representation mismatch between transition instructions and completed visual states. We construct a synthetic text-only evaluation set with Qwen3-32B.

\subsubsection{Controlled Text-Only Evaluation}

Specifically, we ask Qwen3-32B to synthesize 4,096 structured six-tuples. Each tuple consists of an operation type, a static reference description, a dynamic modification instruction, a static target description, a paraphrased target-side static query, and a set of hard negative descriptions. The operation types cover eight representative dynamic semantic categories: addition, removal, replacement, attribute change, relation change, count change, style change, and background change. Each category contains 512 samples. For each sample, we compare two CLIP-based retrieval settings, hereafter termed the target-side static query formulation and the dynamic composed query formulation. In the static setting, the query is expressed as a static target description. In the dynamic setting, the exact template used by the implementation is ``\texttt{REFERENCE, but INSTRUCTION}'', where \texttt{REFERENCE} and \texttt{INSTRUCTION} are replaced by the static reference description and the lowercased dynamic instruction, respectively. Both queries are encoded by the same CLIP text encoder and used to retrieve the corresponding target from the target and hard-negative candidates. Therefore, the only difference between the two settings is whether the query is represented as static semantics of the target or dynamic modification semantics. For conciseness, we refer to these settings below as the static and dynamic formulations, respectively.

For implementation, we evaluate three OpenCLIP text encoders, including ViT-B/32, ViT-L/14, and ViT-G/14, with LAION pretrained weights. All reference descriptions, dynamic queries, static queries describing the target, target descriptions, and hard negatives are encoded by the corresponding CLIP text encoder and $\ell_2$ normalized before similarity computation. For each synthesized sample, the retrieval gallery consists of the ground-truth target description and its hard negatives. We report R@1 by checking whether the query retrieves the corresponding target description as the nearest neighbor under cosine similarity. This design isolates the semantic behavior of the text encoder of CLIP and removes the influence of image encoding or dataset-specific visual noise.

\begin{table}[t]
    \centering
    \caption{
    Controlled text-only retrieval results for target-side static and dynamic composed query formulations under different OpenCLIP backbones. Gap denotes the R@1 difference between the two formulations.
    }
    \label{tab:clip_dynamic_backbone}
    \small
    \setlength{\tabcolsep}{6pt}
    \begin{tabular}{@{}lccc@{}}
        \toprule
        Backbone & Static R@1 & Dynamic R@1 & Gap \\
        \midrule
        ViT-B/32 & 81.52 & 29.69 & 51.83 \\
        ViT-L/14 & 83.23 & 29.00 & 54.22 \\
        ViT-G/14 & 86.21 & 32.59 & 53.61 \\
        \bottomrule
    \end{tabular}
\end{table}

\begin{figure}[t]
    \centering
    \includegraphics[width=0.92\linewidth]{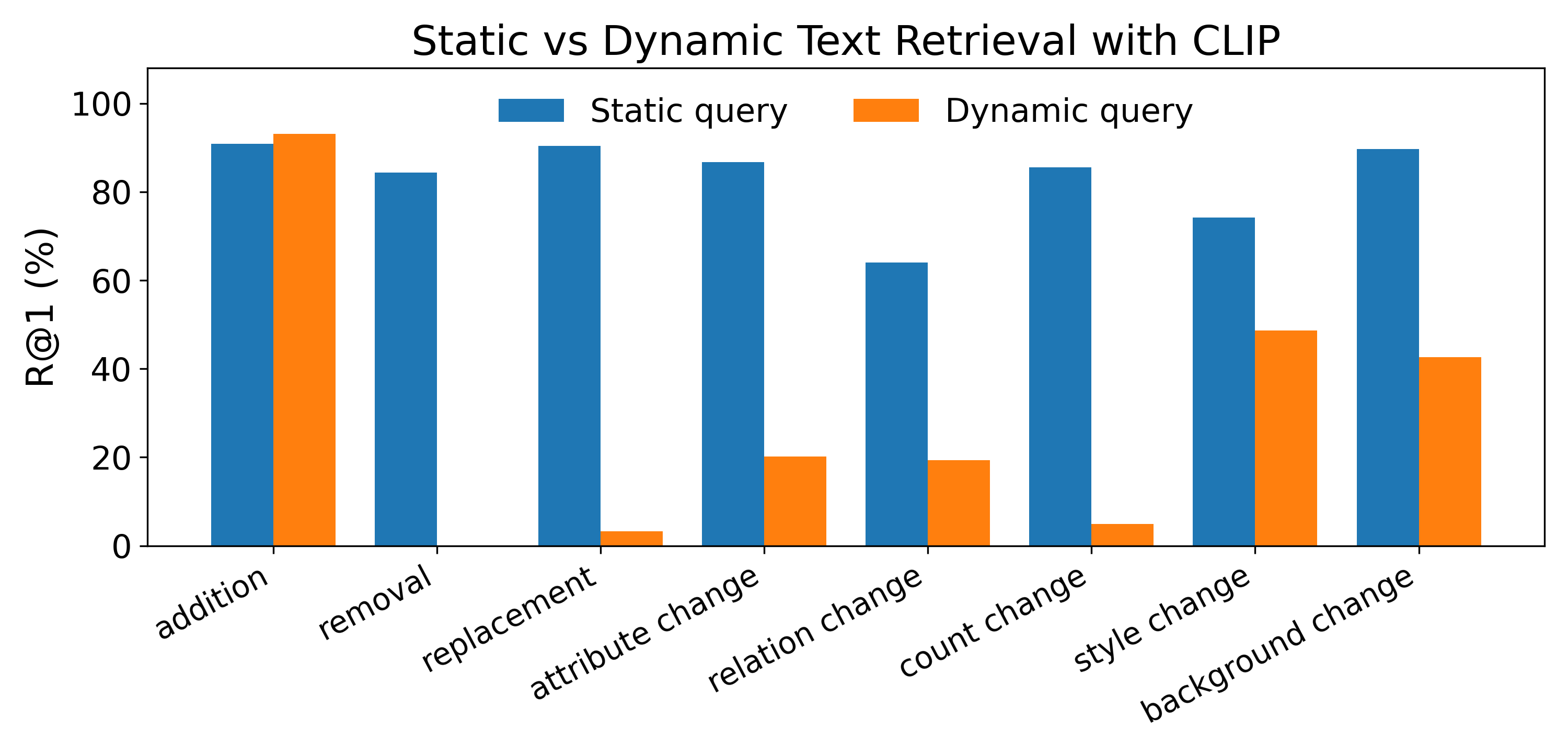}
    \caption{
    R@1 comparison between target-side static and dynamic composed query formulations under the default OpenCLIP ViT-L/14 text encoder. The synthetic evaluation set contains 4,096 Qwen3-32B generated samples across eight dynamic semantic categories.
    }
    \label{fig:clip_dynamic_r1}
\end{figure}

Table~\ref{tab:clip_dynamic_backbone} summarizes the overall results across the three backbones. The static query formulation consistently achieves much higher R@1 than the dynamic formulation. With ViT-B/32, static queries obtain 81.52\% R@1, while dynamic queries only obtain 29.69\%. With the default ViT-L/14 backbone, the gap increases to 54.22 points, with 83.23\% R@1 for static queries and 29.00\% R@1 for dynamic queries. The same phenomenon remains under the stronger ViT-G/14 backbone, where static queries reach 86.21\% R@1 but dynamic queries still remain at 32.59\%. These results suggest that simply scaling the CLIP backbone improves static semantic matching, but does not eliminate the gap between static and dynamic query formulations under this controlled protocol.

\subsubsection{Operation-Wise Analysis}
Fig.~\ref{fig:clip_dynamic_r1} further shows the results for individual operations under the default ViT-L/14 text encoder. The degradation is particularly severe for removal, replacement, and count change, where the dynamic formulation obtains much lower R@1 than the static formulation. For example, dynamic queries obtain 0.00\% R@1 on removal, 3.32\% on replacement, and 4.88\% on count change. Similar trends are also observed with ViT-B/32 and ViT-G/14. Addition is a notable exception: under ViT-L/14 and ViT-G/14, the dynamic formulation reaches 93.16\% and 95.51\% R@1, slightly exceeding the corresponding static results of 90.82\% and 93.95\%. This behavior is plausible because an added object can often be matched as an ordinary visible noun phrase without resolving negation or replacing a source-state concept. We provide the complete visualizations for individual operations and feature distributions for ViT-B/32 and ViT-G/14 in the appendix. Overall, the results indicate that, in this synthetic text-only setting, directly encoding instructions combining the reference and modification is less reliable for operations that require suppressing or transforming source-state semantics, while simple addition can remain well aligned with CLIP's text space.

\begin{figure}[t]
    \centering
    \includegraphics[width=0.86\linewidth]{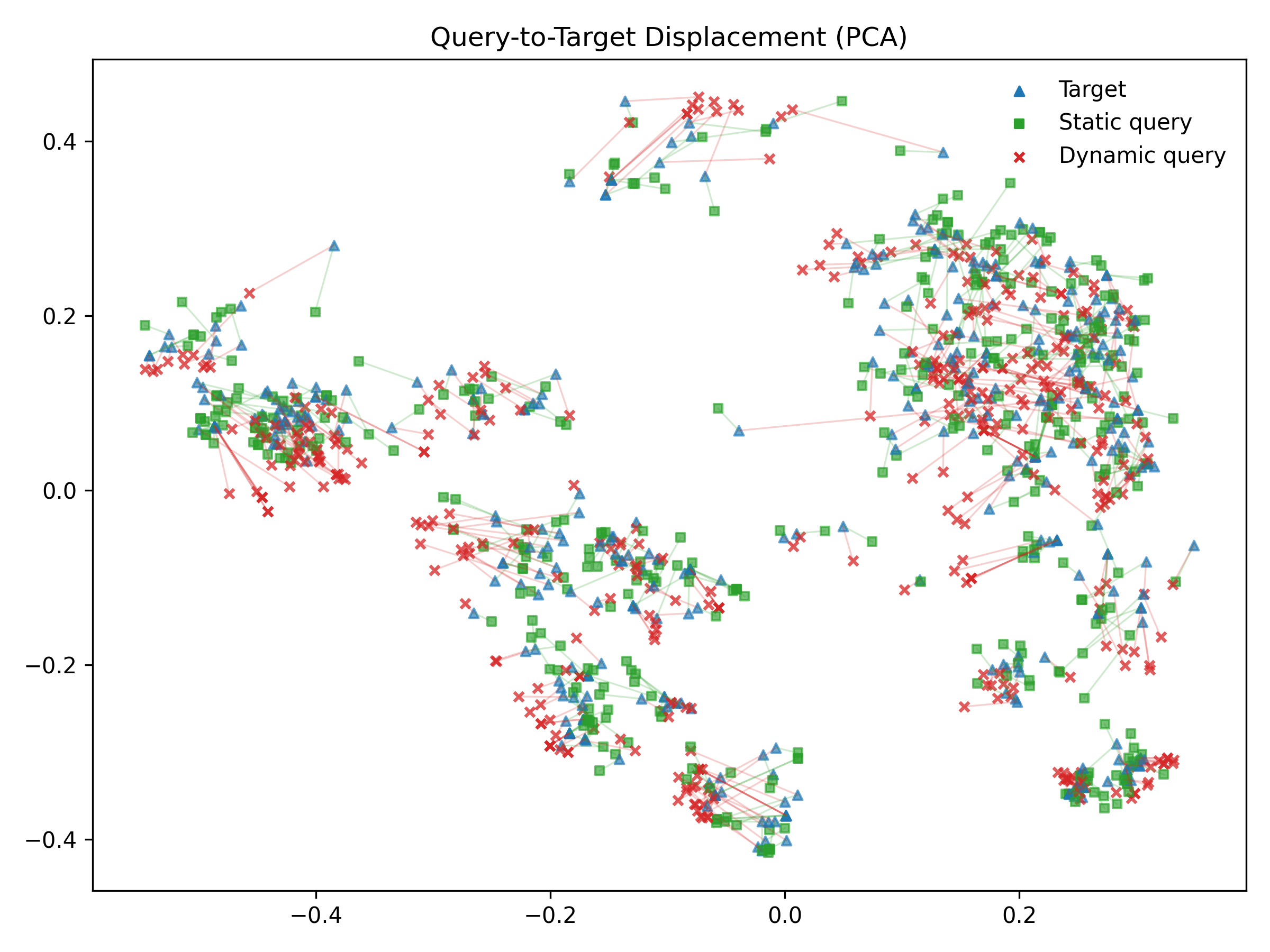}
    \caption{
    Target-anchored principal component analysis (PCA) projection of CLIP text features under the default OpenCLIP ViT-L/14 text encoder. For each of the eight operation types, 50 query samples are randomly selected. Squares denote target-side static query features, crosses denote dynamic composed query features, and triangles denote target features.
    }
    \label{fig:clip_dynamic_feature_visualization}
\end{figure}

\subsubsection{Feature-Space Analysis}
To further inspect the feature distribution, we randomly sample 50 queries from each of the eight operation types and visualize their default ViT-L/14 text features, as shown in Fig.~\ref{fig:clip_dynamic_feature_visualization}. The static query features are generally closer to the corresponding target features, while the dynamic query features are more frequently displaced away from the target semantics. Combined with the consistent R@1 gaps across ViT-B/32, ViT-L/14, and ViT-G/14, this observation provides feature-level evidence that target-state formulations are better aligned with target descriptions than transformation-oriented formulations in our diagnostic setting. This supports the central design choice of ASAP-CIR: reconstruct the state to be retrieved before entering the pretrained vision-language space.

\subsection{Understanding the Multi-Level Target Representation}
\label{subsec:ablation_study}

To determine where the gains of ASAP-CIR originate, we conduct ablation studies on CIRCO using ViT-L/14 as the backbone. CIRCO is selected because its multiple ground-truth targets provide a fine-grained test of whether candidates preserve the intended target state rather than merely one textual realization. We examine the complementary roles of holistic and atomic evidence, importance weighting of atomic semantics, robustness to the choice of MLLM parser, and the calibration and aggregation of atomic evidence.

\begin{table*}[t]
    \centering
    \caption{Component ablation of holistic state alignment, atomic constraint grounding, and importance weighting on CIRCO with ViT-L/14.}
    \label{tab:global_local_component_ablation}
    \footnotesize
    \setlength{\tabcolsep}{5.0pt}
    \begin{tabular}{@{}lccccc@{}}
        \toprule
        Matching evidence & \shortstack{Importance\\weighting} & mAP@5 & mAP@10 & mAP@25 & mAP@50 \\
        \midrule
        Holistic only & -- & 35.19 & 36.14 & 38.88 & 40.03 \\
        Atomic only & $\times$ & 11.96 & 12.61 & 13.92 & 14.66 \\
        Atomic only & $\checkmark$ & 13.19 & 13.84 & 15.26 & 16.07 \\
        Holistic + atomic & $\times$ & 35.40 & 36.39 & 39.10 & 40.25 \\
        \rowcolor{black!7}\textbf{Holistic + atomic} & $\checkmark$ & \textbf{35.49} & \textbf{36.46} & \textbf{39.21} & \textbf{40.35} \\
        \bottomrule
    \end{tabular}
\end{table*}

\subsubsection{Contribution of Holistic and Atomic Evidence}
Table~\ref{tab:global_local_component_ablation} jointly examines holistic alignment, atomic grounding, and semantic importance weighting. Holistic evidence alone is substantially stronger than either atomic-only variant, confirming that complete target descriptions provide the principal retrieval signal. Nevertheless, importance weighting clearly improves the atomic representation itself: compared with uniform weighting, it raises atomic-only mAP@5/10/25/50 from 11.96/12.61/13.92/14.66 to 13.19/13.84/15.26/16.07, corresponding to gains of 1.23, 1.23, 1.34, and 1.41 points. When both evidence sources are aggregated, replacing predicted importance scores with uniform weights yields 35.40/36.39/39.10/40.25, whereas the complete model reaches 35.49/36.46/39.21/40.35. The smaller gains of 0.09, 0.07, 0.11, and 0.10 points in the complete setting are expected because atomic evidence receives only 0.05 of the final score weight in the optimal CIRCO configuration. These results show that holistic semantics establish the target state, while importance-weighted atomic semantics provide complementary fine-grained evidence.

\subsubsection{Robustness to the MLLM Parser}
\begin{table*}[t]
    \centering
    \caption{Robustness to the choice of MLLM parser on CIRCO. Both variants use the ViT-L/14 retrieval backbone and identical retrieval settings.}
    \label{tab:mllm_parser_robustness}
    \footnotesize
    \setlength{\tabcolsep}{6pt}
    \begin{tabular}{@{}llcccc@{}}
        \toprule
        Variant & MLLM parser & mAP@5 & mAP@10 & mAP@25 & mAP@50 \\
        \midrule
        \textbf{ASAP-CIR} & Qwen3.6-27B & \textbf{35.49} & \textbf{36.46} & \textbf{39.21} & \textbf{40.35} \\
        ASAP-CIR* & InternVL3.5-38B & 29.42 & 30.03 & 32.67 & 33.69 \\
        \bottomrule
    \end{tabular}
\end{table*}

We assess whether the proposed target-state formulation depends on a particular MLLM by replacing Qwen3.6-27B with InternVL3.5-38B while keeping the prompts, ViT-L/14 retrieval backbone, candidate features, and all retrieval hyperparameters fixed. The resulting ASAP-CIR* variant obtains 29.42, 30.03, 32.67, and 33.69 mAP at $k=\{5,10,25,50\}$, as reported in Table~\ref{tab:mllm_parser_robustness}. These values are 6.07, 6.43, 6.54, and 6.66 points below the default parser, but remain 3.87, 3.76, 3.86, and 3.70 points above STiTch, respectively.

Inspection of the generated semantics explains part of this gap. InternVL occasionally produces hallucination-like source-state carryover or omits a decisive modification constraint. In CIRCO query ID 638, the instruction requests a teddy bear inside a toy box instead of on a shelf, but the reconstructed target remains a teddy bear on a bookshelf. In query ID 678, the requirement that a dog has only two paws on the ground is omitted entirely. Such traceable errors introduce incorrect or incomplete visual evidence before retrieval and consequently propagate to the final ranking. Nevertheless, the InternVL variant remains competitive across all four metrics, indicating that the representation and matching pipeline is not tied to Qwen alone. The comparison therefore supports cross-MLLM robustness while also showing that accurate target-state reconstruction remains an important performance bottleneck.

\begin{figure}[t]
    \centering
    \begin{minipage}{0.49\linewidth}
        \centering
        \includegraphics[width=\linewidth]{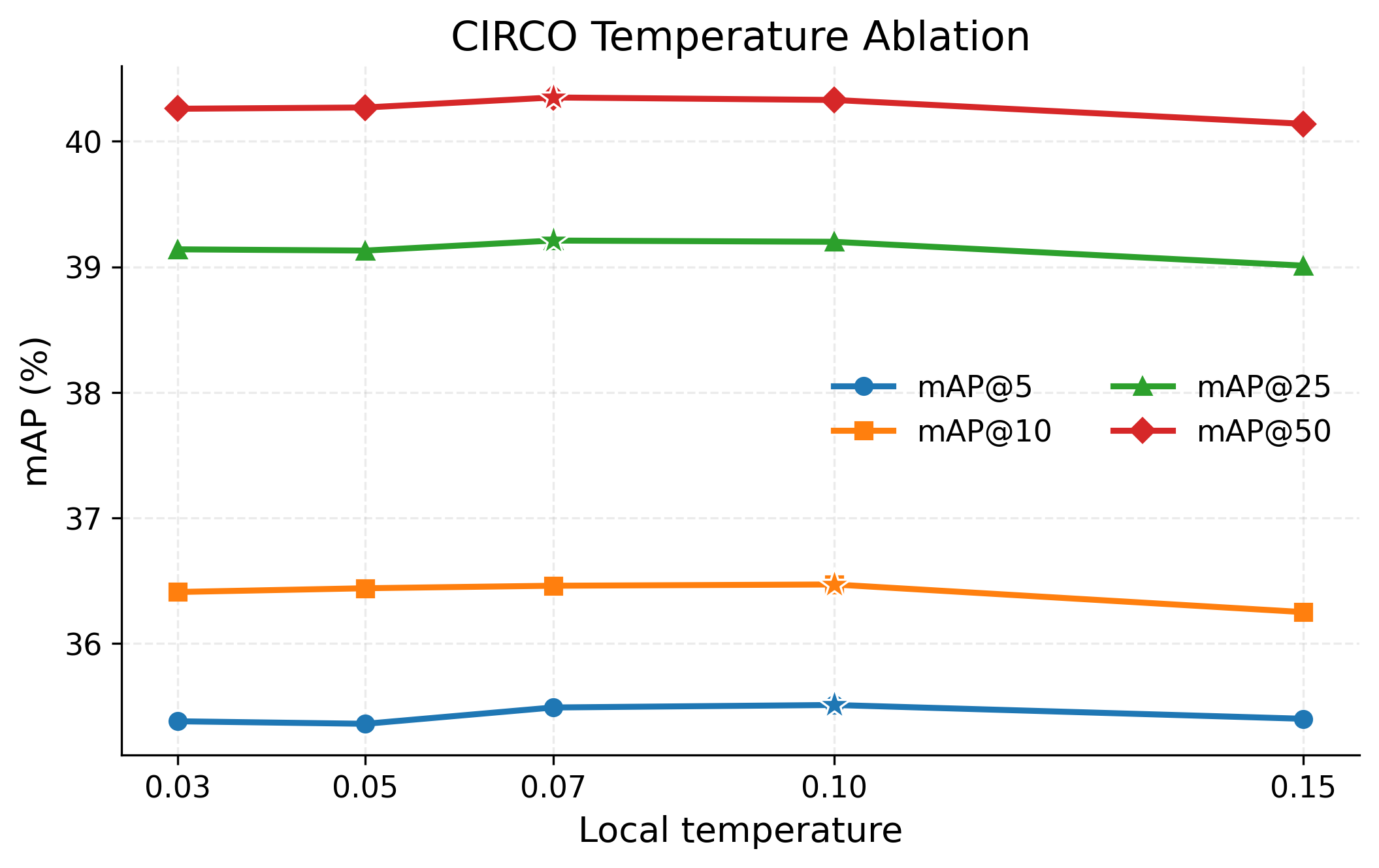}
        \vspace{0.2em}
        \centerline{(a) Temperature}
    \end{minipage}
    \hfill
    \begin{minipage}{0.49\linewidth}
        \centering
        \includegraphics[width=\linewidth]{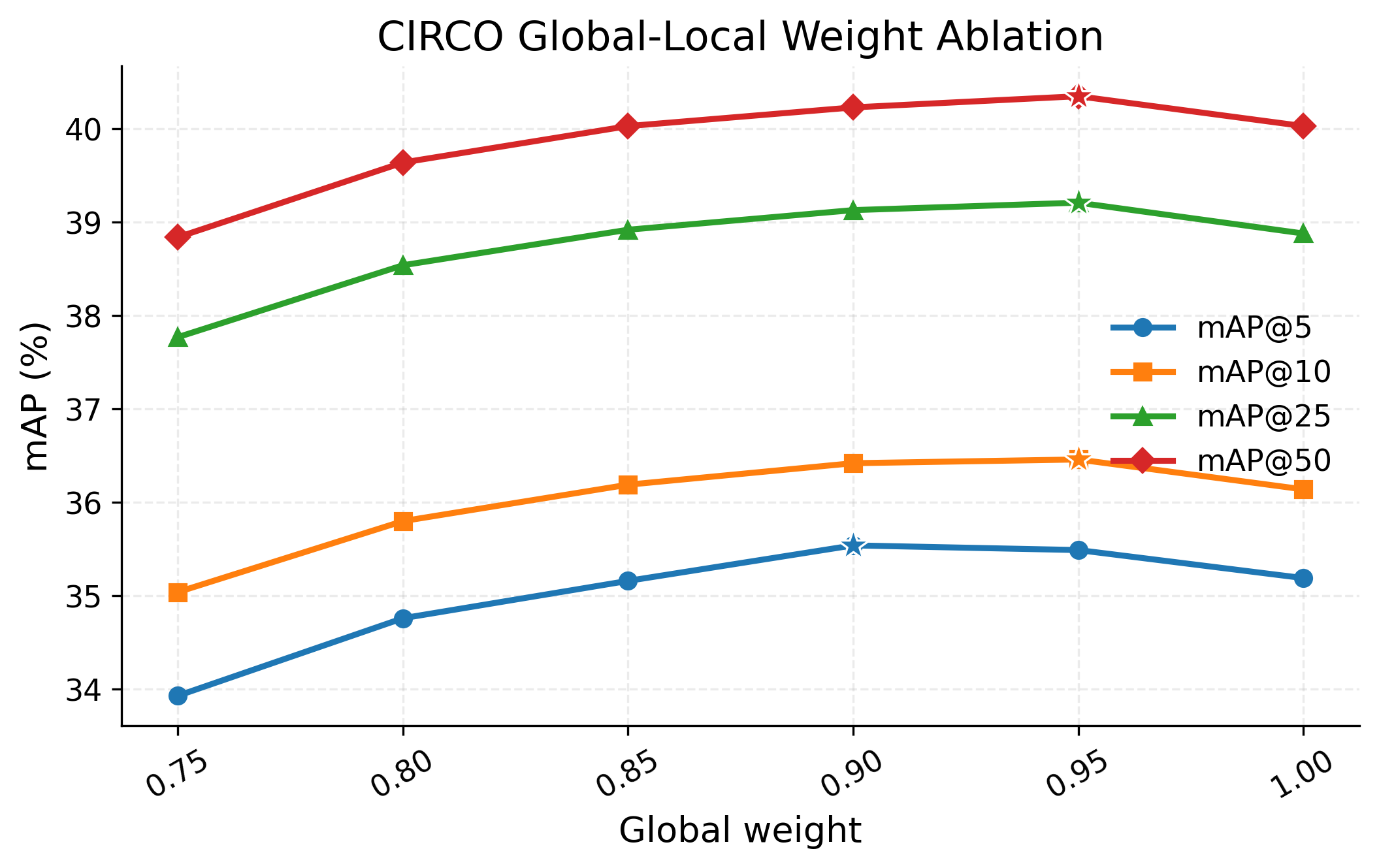}
        \vspace{0.2em}
        \centerline{(b) Fusion weights}
    \end{minipage}
    \caption{
    Ablation studies on CIRCO. (a) Effect of the atomic grounding temperature with holistic/atomic weights fixed to $0.95/0.05$. (b) Effect of holistic/atomic fusion weights with the grounding temperature fixed to $\tau=0.07$.
    }
    \label{fig:circo_ablation}
\end{figure}

\subsubsection{Temperature of Atomic Evidence Grounding}
The atomic constraint grounding module computes the soft semantic coverage score of each atomic semantic by applying a softmax operation over candidate patch features. The temperature $\tau$ controls the sharpness of this atomic-to-patch assignment. As shown in Fig.~\ref{fig:circo_ablation}(a), ASAP-CIR is relatively insensitive to the temperature within a reasonable range. When $\tau$ varies from 0.03 to 0.15, all four mAP metrics remain stable, suggesting that atomic constraint grounding does not rely on a carefully tuned temperature. The best or near-best results are obtained around $\tau=0.07$ and $\tau=0.10$. In our main experiments, we set $\tau=0.07$ because it provides consistently strong performance across different retrieval depths while keeping the atomic-to-patch assignment sufficiently selective.

\subsubsection{Target-State Evidence Aggregation}
We then evaluate the contribution of holistic and atomic evidence by varying the holistic weight $\lambda_g$ from 0.75 to 1.00, with the atomic weight set to $\lambda_l=1-\lambda_g$. As shown in Fig.~\ref{fig:circo_ablation}(b), holistic evidence alone already achieves competitive performance, suggesting that the holistic descriptions generated by the MLLM provide a strong static representation of the composed query. However, introducing a small amount of atomic evidence consistently improves ranking quality. In particular, the setting $\lambda_g=0.95$ and $\lambda_l=0.05$ achieves the best performance on mAP@10, mAP@25, and mAP@50, while $\lambda_g=0.90$ and $\lambda_l=0.10$ gives the best mAP@5. This suggests that importance-weighted atomic semantics complement holistic descriptions by emphasizing fine-grained requirements that may be weakened in a single overall caption. Meanwhile, assigning an overly large atomic weight degrades performance, since atomic semantics alone cannot fully represent the overall target appearance. Therefore, we use $\lambda_g=0.95$ and $\lambda_l=0.05$ as the default fusion weights for CIRCO.

\subsection{Qualitative Analysis and Limitations}
\label{subsec:qualitative_results}

\begin{figure*}[t]
    \centering
    \includegraphics[width=0.95\textwidth]{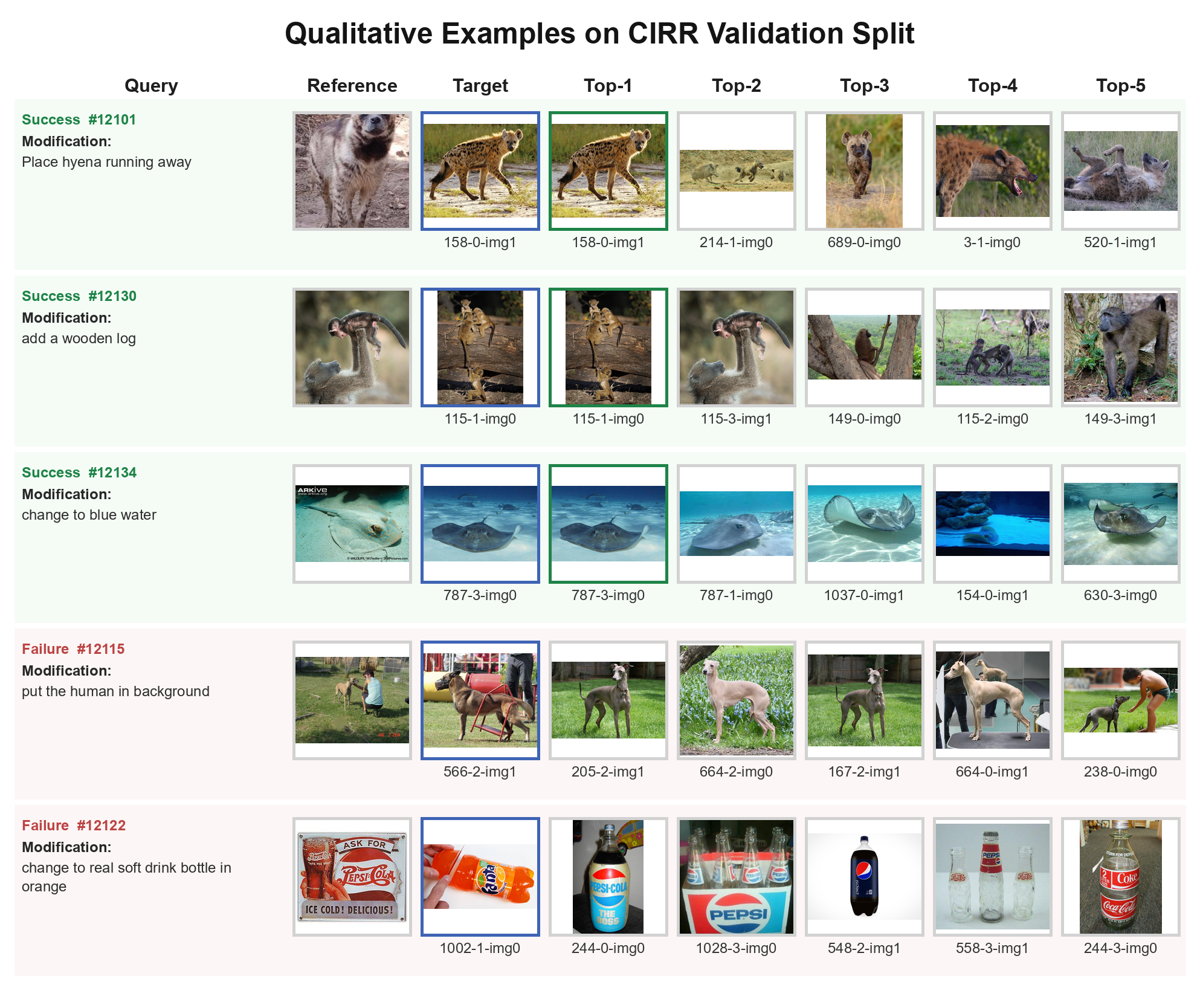}
    \caption{
    Qualitative examples on the CIRR validation split. Each row shows the reference image, the ground-truth target image, and the top-5 retrieved candidates. The first three rows are successful cases and the final two rows are failure cases. A green border marks the correctly retrieved target
    }
    \label{fig:cirr_val_qualitative_examples}
\end{figure*}

We further provide qualitative examples to intuitively analyze the retrieval behavior of ASAP-CIR. Since the official CIRR test split does not provide target images for visualization, we conduct this qualitative analysis on the CIRR validation split, where the ground-truth target image is available. Fig.~\ref{fig:cirr_val_qualitative_examples} shows five representative examples, including three successful retrieval cases and two failure cases. For each query, we present the reference image, the ground-truth target image, and the top-5 retrieved candidates. A green border indicates that the retrieved candidate matches the ground-truth target.

\subsubsection{Successful Target-State Reconstruction}
The first successful case requires the hyena to be running away. The reconstructed holistic state preserves the animal and scene context, while the atomic action constraint distinguishes the target motion from visually similar animal images. In the second case, the query introduces a wooden log into the reference scene. Holistic alignment maintains the overall scene content, whereas atomic grounding emphasizes the newly added object and suppresses candidates that preserve only the reference appearance. The third case changes the water to blue. Here, the holistic description retains the surrounding scene, while the explicit color constraint helps separate the target from candidates with similar content but a different water appearance. These cases illustrate the complementary roles of holistic state alignment and atomic constraint grounding under different modification types.

\subsubsection{Failure Cases and Remaining Boundaries}
The failure cases expose the distinction between reconstructing a target state and grounding it precisely. In the example ``put the human in background'', the parser can express the desired entities, yet the retriever fails to resolve their spatial relation. In the example ``change to real soft drink bottle in orange'', the top-ranked candidates contain relevant soft drink bottles but do not match the expected orange instance. ASAP-CIR may therefore remain unreliable for precise relations, subtle attributes, or instance-level distinctions among visually similar candidates. Target-state reconstruction improves the query representation, but it does not by itself solve every visual-grounding ambiguity.

\section{Discussion}
\label{sec:discussion}

The results support a representation-centered interpretation of ZS-CIR. A composed query naturally describes a transition, whereas a retrieval candidate depicts a completed state. ASAP-CIR reduces this asymmetry by reconstructing the expected state before matching, allowing the pretrained vision-language model to operate closer to its image--caption alignment regime. The resulting representation is deliberately multi-level: holistic descriptions retain overall coherence, while atomic semantics preserve requirements that can be diluted in a single caption.

The diagnostic results also clarify when this principle is most useful. The largest static--dynamic gaps occur for removal, replacement, and count change, where successful retrieval requires suppressing or transforming source-state concepts. Addition is less dependent on reconstruction because the added concept can often be matched as an ordinary visible noun phrase. Thus, the benefit is operation dependent rather than universal, and the controlled analysis should be interpreted as evidence for the formulation under the studied protocol rather than as a general claim about every CLIP capability.

Finally, semantic reconstruction and visual grounding remain distinct problems. An MLLM may correctly describe the target state while CLIP patch features still fail to localize a subtle relation, distinguish a specific instance, or resolve visually ambiguous attributes. Parser errors can likewise propagate into retrieval. Future work can therefore build on target-state reconstruction while improving uncertainty modeling, relation-aware grounding, and parser--retriever interaction without requiring supervised CIR triplets.

\section{Conclusion}
\label{sec:conclusion}

In this paper, we studied how composed retrieval intent should be represented for pretrained vision-language retrieval. Our controlled analysis shows that, under the considered text-only protocol, target-state descriptions are more reliably aligned with intended targets than formulations that combine a reference description and modification instruction. We proposed ASAP-CIR, a training-free realization of target-state reconstruction that combines holistic descriptions with importance-weighted atomic semantics, followed by holistic state alignment, atomic constraint grounding, and calibrated target-state evidence aggregation. Experiments on FashionIQ, CIRR, and CIRCO demonstrate the utility of this formulation, with its clearest gains on CIRCO. The results also expose important boundaries: performance remains sensitive to parser quality, complex spatial relations, subtle instance-level grounding, and dataset-specific retrieval structure. We hope this representation-centered view encourages future work to examine not only how multimodal signals are composed, but also how user intent should be formulated before it enters a pretrained vision-language retrieval space.

\backmatter

\bmhead{Statements and Declarations}

\bmhead{Funding}
This work was supported by the Beijing Natural Science Foundation (No.~L259051) and the Natural Science Foundation of Hebei Province (No.~F2025105018).

\bmhead{Competing interests}
The authors have no competing interests to declare that are relevant to the content of this article.

\bmhead{Author contributions}
Yihe Zhao conceived and implemented the method, conducted the experiments, analyzed the results, and drafted the manuscript. Songhe Feng supervised the research and reviewed and edited the manuscript. Both authors read and approved the final manuscript.

\bmhead{Data availability}
The datasets analyzed in this study are publicly available from their official project repositories. Access information for FashionIQ, CIRR, and CIRCO is provided by \citet{Wu_2021_CVPR}, \citet{Liu_2021_ICCV}, and \citet{Baldrati_2023_ICCV}, respectively. No new benchmark dataset was created as part of this study.

\bmhead{Code availability}
The source code will be made publicly available upon acceptance.
\setcounter{mainfigurecount}{\value{figure}}
\begin{appendices}

\section{Dataset-Specific MLLM Prompts}
\label{app:prompts}

\noindent
This appendix provides the query MLLM prompts used by ASAP-CIR. The prompts are directly taken from the implementation. Following the prompt presentation style of recent training-free ZS-CIR works, we show each prompt as a boxed block with a short functional description. Each dataset shares the same compact output instruction and then uses a dataset-specific task prompt.

\prompttitle{Prompt A: Shared Output Format Instruction}
\promptnote{Function.}{This shared instruction standardizes all MLLM responses into holistic descriptions and importance-weighted atomic semantics. The prompt's D-line ``global caption'' field corresponds to the holistic descriptions used in the paper.}
\begin{lstlisting}
Return compact retrieval content only. Every output sentence will be encoded by CLIP, so it must look like a natural web image caption.
Use exactly this line format:
D1: one CLIP-style global caption
D2: one CLIP-style global caption
D3: one CLIP-style global caption
S: scope | one atomic CLIP-style visual phrase or caption | importance
Every S line must contain exactly two vertical bars and three fields: scope, phrase, importance.
Never output an S line without an importance score.
Allowed S scopes are exactly: global, atomic.
Use global only for one complete caption summary; use atomic for every local object, attribute, relation, scene, style, color, count, state, or material phrase.
Importance is a number from 0.20 to 1.00.
Use numeric importance with two decimals, for example 1.00, 0.85, 0.60, or 0.40.
Do not output JSON, markdown, explanations, reasoning, bullet markers, or extra text.
D lines should usually start with caption phrases such as 'A photo of', 'A close-up photo of', 'A product photo of', or 'An image of'.
S lines should be short caption noun phrases or simple visual captions, not analytical sentences.
Avoid meta phrases such as 'the target image shows', 'the target image', 'the image shows', 'visible in the image', 'main subject', 'query', or 'retrieval'.
Do not mention the reference image, source image, original image, modification text, or editing operation.
Forbidden output words and phrases: target image, reference, original, source, modify, modified, edit, edited, add, added, remove, removed, replace, replaced, change, changed, become, instead, compared with, different from, same as, remain, remains, unchanged, no longer, without.
If the modification implies absence, write only positive visible target facts, not negative facts.
If the modification implies identity or size preservation, describe the visible target size/shape directly, not its relation to another image.
Prefer fewer high-confidence S lines over many weak S lines.
\end{lstlisting}

\prompttitle{Prompt B: FashionIQ Static Semantic Parsing for Queries}
\promptnote{Function.}{This prompt focuses the MLLM on fashion-specific target attributes such as garment category, color, material, texture, pattern, fit, silhouette, and style.}
\begin{lstlisting}
Task: FashionIQ composed fashion retrieval.
You receive a reference fashion image and a modification text.
Infer the desired fashion item from both inputs. Use the image to identify visually salient garment details, but write only standalone CLIP-style captions.
Focus on clothing category, color, material, texture, pattern, sleeve length, neckline, fit, silhouette, decorations, and style.
Assign high importance to fashion details related to the modification, medium importance to preserved salient garment details, and low importance to unrelated pose/background.
All D and S lines must be natural caption text. Do not write relative phrases such as "different from the reference" or "same as the reference".
Produce 3 D lines and 4 to 7 S lines.

Example:
Modification text: "is shiny and silver with shorter sleeves and fit and flare"
D1: A product photo of a shiny silver fit-and-flare {fashion_type} with short sleeves.
D2: A catalog photo of a short-sleeved silver {fashion_type} with a shiny finish and a fit-and-flare silhouette.
D3: A fashion product image of a shiny silver short-sleeved {fashion_type} with a flared shape.
S: global | A shiny silver fit-and-flare {fashion_type} with short sleeves. | 1.00
S: atomic | A {fashion_type}. | 0.80
S: atomic | Shiny silver fabric. | 1.00
S: atomic | Short sleeves. | 0.95
S: atomic | A fit-and-flare silhouette. | 0.95
S: atomic | A polished fashion product photo. | 0.60

Fashion type: {fashion_type}
Modification text: "{modification_text}"
\end{lstlisting}

\prompttitle{Prompt C: CIRR Static Semantic Parsing for Queries}
\promptnote{Function.}{This prompt converts general composed-image modifications into standalone static visual facts of the target, including objects, counts, attributes, relations, scenes, and backgrounds.}
\begin{lstlisting}
Task: CIRR composed image retrieval.
You receive a reference image and a modification text.
Infer the desired final visual content from both inputs. Convert dynamic modifications into static visible caption facts.
Focus on objects, people, animals, object count, attributes, pose, object condition, spatial relations, interactions, scene, background, and composition.
For removals or negative instructions, describe only the visible content that should be present.
Preserve unchanged content only when it is visually important or likely useful for retrieval.
All D and S lines must be natural CLIP-style captions. Do not describe editing operations or compare with the reference image.
Produce 3 D lines and 4 to 7 S lines.

Example:
Modification text: "remove all but one dog and add a woman hugging it"
D1: A photo of a woman hugging one dog.
D2: A natural image of one dog being hugged by a woman.
D3: A scene with a woman holding and hugging a single dog.
S: global | A woman hugging one dog. | 1.00
S: atomic | One dog. | 1.00
S: atomic | A woman. | 1.00
S: atomic | A woman hugging a dog. | 1.00
S: atomic | A central dog. | 0.85
S: atomic | A natural scene around a woman and a dog. | 0.45

Example:
Modification text: "same size of product but different colors"
D1: A product photo of a floral tennis racket case with a HEAD logo and a colorful surface pattern.
D2: A catalog image of a rectangular sports racket case with a multicolor floral design and a brown logo patch.
D3: A photo of a HEAD racket case with a slim rectangular shape, floral decoration, and varied colors.
S: global | A colorful floral HEAD tennis racket case. | 1.00
S: atomic | A HEAD tennis racket case. | 1.00
S: atomic | A slim rectangular racket case. | 0.90
S: atomic | A multicolor floral surface pattern. | 1.00
S: atomic | A brown logo patch on the case. | 0.80
S: atomic | A product on a plain background. | 0.45

Example:
Modification text: "remove the background and show a cow and two dogs"
D1: A photo of one cow and two dogs.
D2: A simple image of one cow with two dogs.
D3: A scene with a cow and two dogs as the dominant animals.
S: global | One cow and two dogs. | 1.00
S: atomic | One cow. | 1.00
S: atomic | Two dogs. | 1.00
S: atomic | A cow together with two dogs. | 0.90
S: atomic | A simple animal scene. | 0.70

Modification text: "{modification_text}"
\end{lstlisting}

\prompttitle{Prompt D: CIRCO Static Semantic Parsing for Queries}
\promptnote{Function.}{This prompt preserves the shared concept while converting the modification text into CLIP-style holistic descriptions and importance-weighted atomic semantics.}
\begin{lstlisting}
Task: CIRCO composed image retrieval.
You receive a reference image, a shared concept, and a modification text.
Infer the desired final visual content by preserving the shared concept and satisfying the modification. Use the image as visual evidence, but write only standalone CLIP-style captions.
Focus on shared concept, objects, people, attributes, relations, lighting, viewpoint, image quality, time of day, scene, and COCO-style natural image content.
Assign high importance to the shared concept and facts related to the modification.
All D and S lines must be natural caption text. Do not mention reference/source/original images or editing operations.
Produce 3 D lines and 4 to 7 S lines.

Example:
Shared concept: "a man sitting on an outdoor toilet"
Modification text: "has a higher quality and is taken during the daytime"
D1: A high-quality daytime photo of a man sitting on an outdoor toilet.
D2: A clear outdoor daytime photo of a man sitting on an outdoor toilet.
D3: A natural image of a man sitting on an outdoor toilet during the day.
S: global | A high-quality daytime scene of a man sitting on an outdoor toilet. | 1.00
S: atomic | A man. | 1.00
S: atomic | An outdoor toilet. | 0.95
S: atomic | A man sitting on an outdoor toilet. | 1.00
S: atomic | An outdoor daytime scene. | 1.00
S: atomic | A high-quality natural photo. | 0.95

Shared concept: "{shared_concept}"
Modification text: "{modification_text}"
\end{lstlisting}

\section{Backbone Analysis}
\label{app:clip_dynamic_backbone}

\noindent
This appendix provides additional results across backbones for the controlled dynamic semantic analysis in Sec.~\ref{subsec:clip_dynamic_analysis}. In the main paper, we use ViT-L/14 as the representative backbone for detailed operation-wise and feature-level visualization. Here, we further report the corresponding results for ViT-B/32 and ViT-G/14. The evaluation protocol is exactly the same as in Sec.~\ref{subsec:clip_dynamic_analysis}: for each synthesized six-tuple, the target-side static query and the dynamic composed query are encoded by the same CLIP text encoder and retrieve the target description from the gallery containing the target and hard negatives.

\setcounter{figure}{\value{mainfigurecount}}
\renewcommand{\thefigure}{\arabic{figure}}
\renewcommand{\theHfigure}{appendix.\arabic{figure}}

\begin{figure}[!htbp]
    \centering
    \begin{minipage}{\linewidth}
        \centering
        \includegraphics[width=\linewidth]{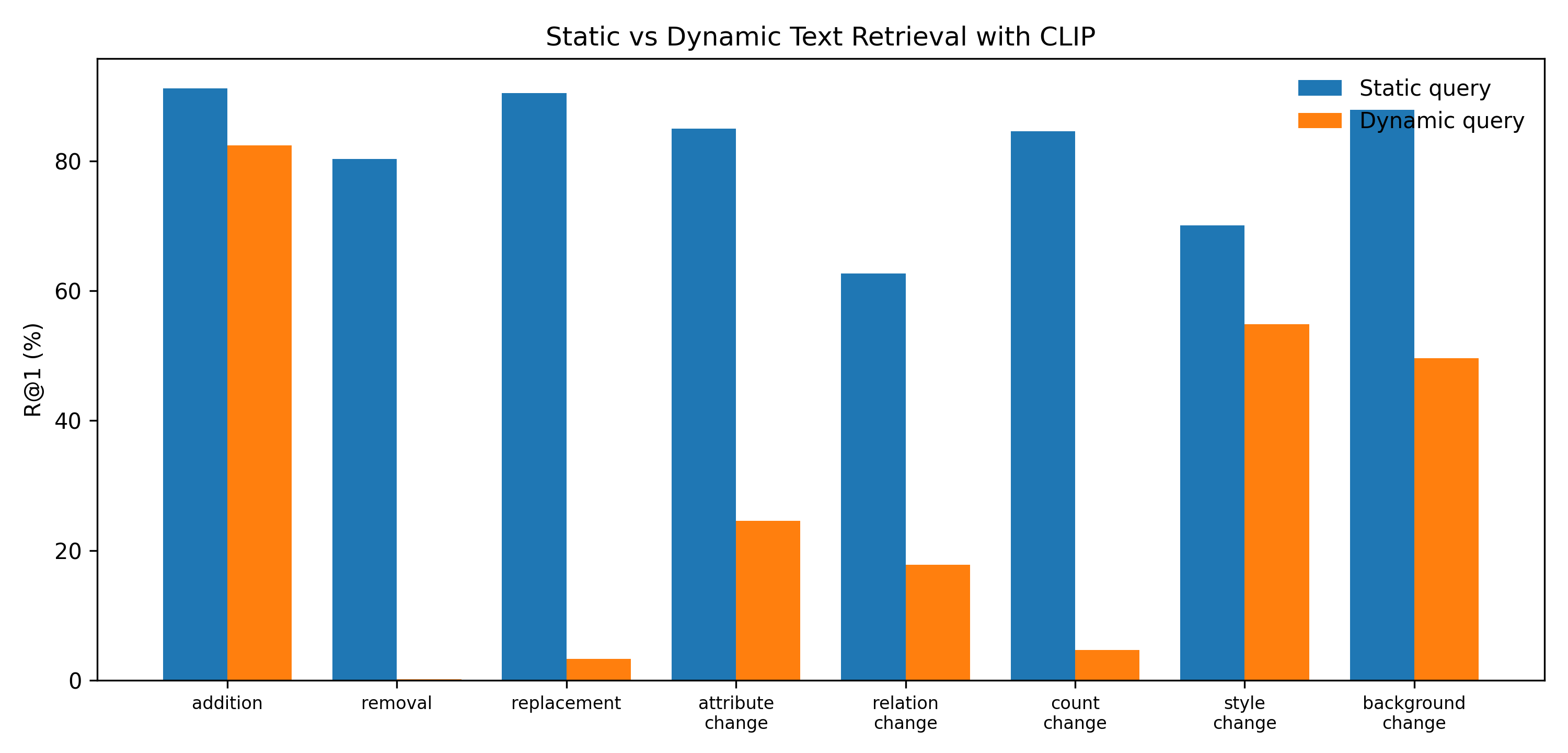}
        \vspace{0.3em}
        \centerline{(a) ViT-B/32}
    \end{minipage}
    \vspace{0.6em}
    \begin{minipage}{\linewidth}
        \centering
        \includegraphics[width=\linewidth]{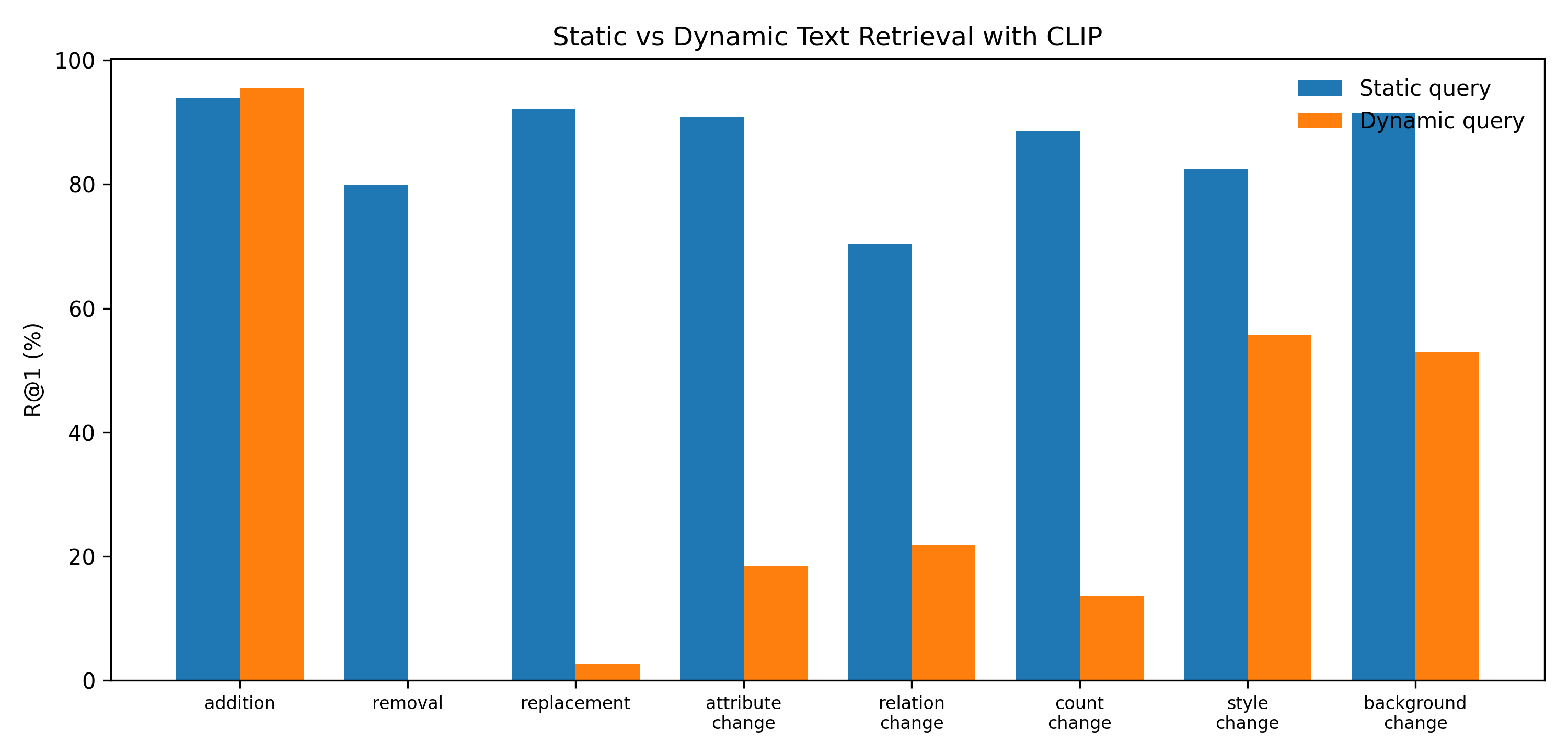}
        \vspace{0.3em}
        \centerline{(b) ViT-G/14}
    \end{minipage}
    \caption{
    Operation-wise R@1 comparison between target-side static and dynamic composed query formulations under additional OpenCLIP backbones.
    }
    \label{fig:appendix_clip_dynamic_operation}
\end{figure}

As shown in Fig.~\ref{fig:appendix_clip_dynamic_operation}, the operation-wise behavior is consistent with the ViT-L/14 results in the main paper. Under ViT-B/32, static queries obtain 81.52\% overall R@1, while dynamic queries obtain only 29.69\%. Under the stronger ViT-G/14, static queries improve to 86.21\%, but dynamic queries still remain at 32.59\%. In both cases, removal, replacement, and count change remain particularly difficult for dynamic query encoding. For example, ViT-B/32 obtains only 0.20\%, 3.32\%, and 4.69\% dynamic R@1 on removal, replacement, and count change, respectively; ViT-G/14 obtains 0.00\%, 2.73\%, and 13.67\% on the same categories. These results further suggest that the observed behavior is not specific to a single CLIP backbone.

\begin{figure}[!htbp]
    \centering
    \begin{minipage}{\linewidth}
        \centering
        \includegraphics[width=\linewidth]{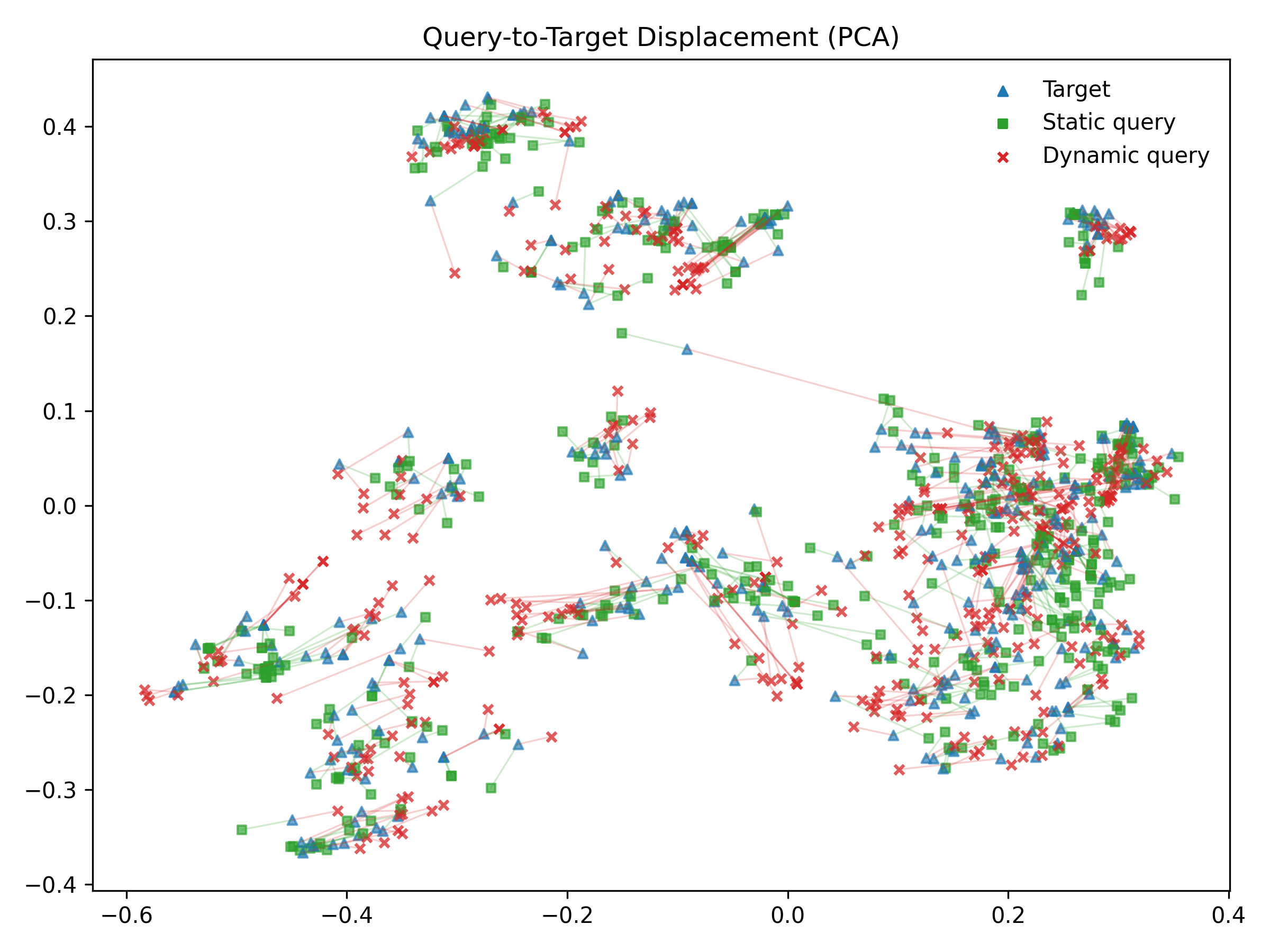}
        \vspace{0.3em}
        \centerline{(a) ViT-B/32}
    \end{minipage}
    \vspace{0.6em}
    \begin{minipage}{\linewidth}
        \centering
        \includegraphics[width=\linewidth]{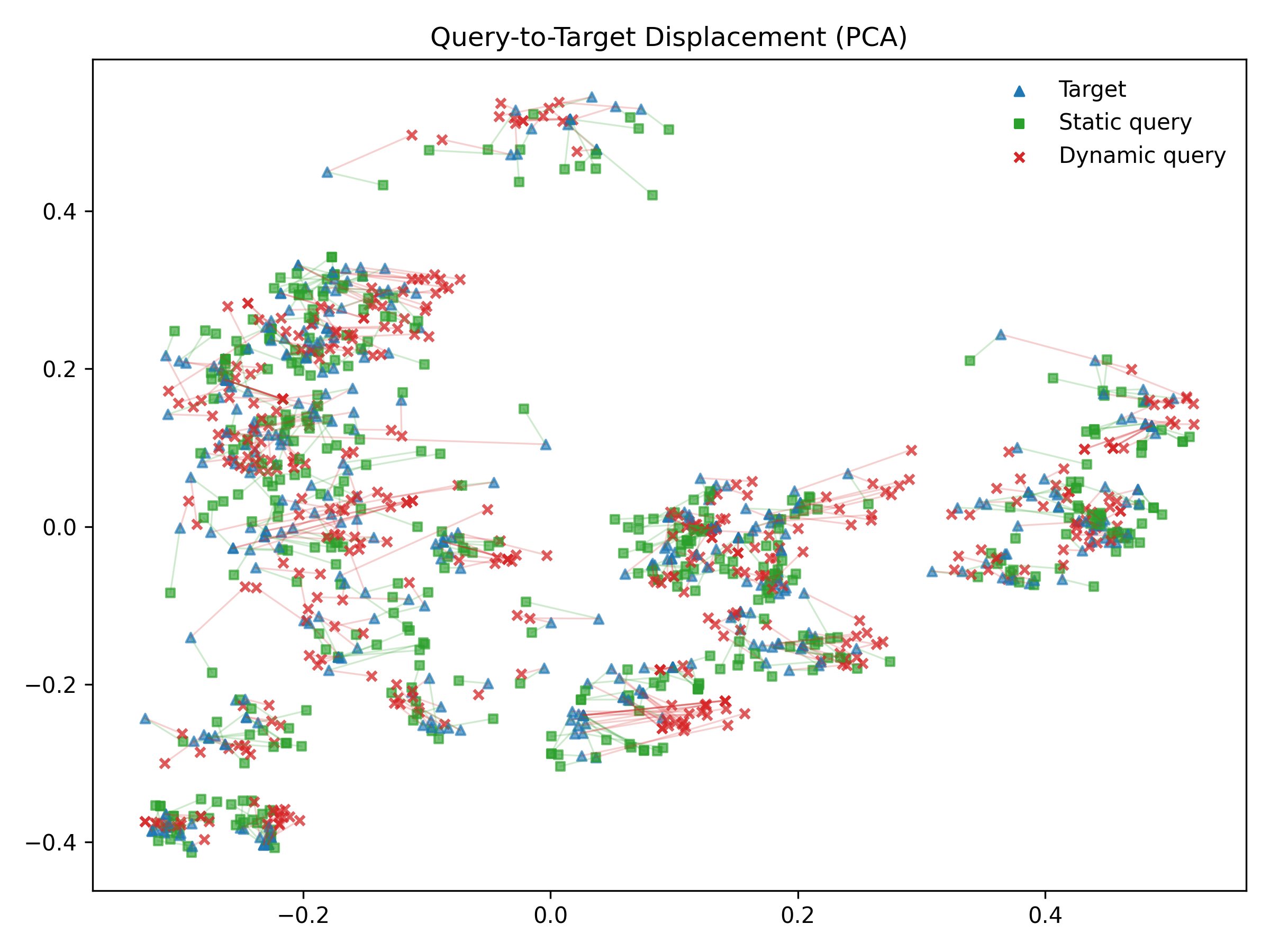}
        \vspace{0.3em}
        \centerline{(b) ViT-G/14}
    \end{minipage}
    \caption{
    Target-anchored PCA projections under additional OpenCLIP backbones. For each operation type, 50 samples are randomly selected. Target-side static query features are generally closer to the target features than dynamic composed query features.
    }
    \label{fig:appendix_clip_dynamic_feature}
\end{figure}

Fig.~\ref{fig:appendix_clip_dynamic_feature} further visualizes the target-anchored feature distributions for ViT-B/32 and ViT-G/14. Although the projected layouts vary across encoders, the qualitative pattern remains stable: static queries describing the target tend to be closer to the corresponding target descriptions, whereas dynamic composed queries are more frequently displaced away from the target semantics. Together with the quantitative results in Table~\ref{tab:clip_dynamic_backbone}, these supplementary results indicate that the observed gap between static and dynamic formulations is persistent across backbone scales. This provides additional evidence for the core motivation of ASAP-CIR: dynamic modification instructions can benefit from being converted into static semantics of the target before CLIP-based matching.

\end{appendices}

\bibliography{references}

\end{document}

%% file: experiment_tables.tex
\begin{table*}[t]
\centering
\caption{FashionIQ validation results. The best and second-best results within each backbone group are shown in bold and underlined, respectively.}
\label{tab:fashioniq_results}
\begingroup
\scriptsize
\setlength{\tabcolsep}{2pt}
\renewcommand{\arraystretch}{1.05}
\resizebox{\textwidth}{!}{%
\begin{tabular}{@{}llcccccccccc@{}}
\toprule
Backbone & Method & Venue & T-Free
& \multicolumn{2}{c}{Dress}
& \multicolumn{2}{c}{Shirt}
& \multicolumn{2}{c}{Toptee}
& \multicolumn{2}{c}{Average} \\
\cmidrule(lr){5-6}\cmidrule(lr){7-8}\cmidrule(lr){9-10}\cmidrule(lr){11-12}
 & & & & R@10 & R@50 & R@10 & R@50 & R@10 & R@50 & R@10 & R@50 \\
\midrule
\multirow{6}{*}{ViT-B/32}
& SEARLE & ICCV'23 & N & 18.54 & 39.51 & 24.44 & 41.61 & 25.70 & 46.46 & 22.89 & 42.53 \\
& FlowCIR & ECCV'26 & N & 24.80 & 41.80 & 19.10 & 40.10 & 26.40 & 47.80 & 23.40 & 43.20 \\
& CIReVL & ICLR'24 & Y & 25.29 & \underline{46.36} & 28.36 & 47.84 & 31.21 & 53.85 & 28.29 & 49.35 \\
& LDRE & SIGIR'24 & Y & 19.97 & 41.84 & 27.38 & 46.27 & 27.07 & 48.78 & 24.81 & 45.63 \\
& AutoCIR & KDD'25 & Y & \underline{26.52} & \underline{46.36} & \textbf{32.43} & \textbf{51.67} & \underline{33.96} & \underline{56.09} & \underline{30.97} & \underline{51.37} \\
& ASAP-CIR (Ours) & -- & Y & \textbf{28.51} & \textbf{52.01} & \underline{29.49} & \underline{49.61} & \textbf{34.98} & \textbf{57.98} & \textbf{30.99} & \textbf{53.20} \\
\midrule
\multirow{11}{*}{ViT-L/14}
& Pic2Word & CVPR'23 & N & 20.00 & 40.20 & 26.20 & 43.60 & 27.90 & 47.40 & 24.70 & 43.73 \\
& SEARLE & ICCV'23 & N & 20.48 & 43.13 & 26.89 & 45.58 & 29.32 & 49.97 & 25.56 & 46.23 \\
& LinCIR & CVPR'24 & N & 20.92 & 42.44 & 29.10 & 46.81 & 28.81 & 50.18 & 26.28 & 46.49 \\
& FlowCIR & ECCV'26 & N & \underline{31.60} & 48.50 & 24.44 & 44.30 & 33.20 & 53.90 & 29.70 & 48.90 \\
& DiffComp & CVPR'26 & N & 27.01 & 49.63 & \textbf{34.79} & \textbf{54.67} & 36.06 & 57.24 & 32.62 & 53.85 \\
& TT-RLDR & AAAI'26 & N & 24.86 & 46.62 & 33.46 & 52.65 & 34.52 & 54.77 & 30.95 & 51.35 \\
& CIReVL & ICLR'24 & Y & 24.79 & 44.76 & 29.49 & 47.40 & 31.36 & 53.65 & 28.55 & 48.57 \\
& LDRE & SIGIR'24 & Y & 22.93 & 46.76 & 31.04 & 51.22 & 31.57 & 53.64 & 28.51 & 50.54 \\
& AutoCIR & KDD'25 & Y & 24.94 & 45.81 & \underline{34.00} & \underline{53.43} & 33.10 & 55.58 & 30.68 & 51.60 \\
& OSrCIR & CVPR'25 & Y & 29.70 & \underline{51.81} & 33.17 & 52.03 & \underline{36.92} & \underline{59.27} & \underline{33.26} & \underline{54.37} \\
& ASAP-CIR (Ours) & -- & Y & \textbf{32.42} & \textbf{52.85} & 33.46 & 51.37 & \textbf{39.11} & \textbf{60.84} & \textbf{35.00} & \textbf{55.02} \\
\bottomrule
\end{tabular}%
}
\endgroup
\end{table*}

\begin{table*}[t]
\centering
\caption{CIRR test1 results. The best and second-best results within each backbone group are shown in bold and underlined, respectively.}
\label{tab:cirr_results}
\begingroup
\scriptsize
\setlength{\tabcolsep}{5pt}
\renewcommand{\arraystretch}{1.08}
\begin{tabular}{@{}llcccccc@{}}
\toprule
Backbone & Method & Venue & T-Free & R@1 & R@5 & R@10 & R@50 \\
\midrule
\multirow{8}{*}{ViT-B/32}
& SEARLE & ICCV'23 & N & 24.00 & 53.42 & 66.82 & 89.78 \\
& TT-RLDR & AAAI'26 & N & 27.91 & -- & \underline{71.98} & \textbf{91.89} \\
& FlowCIR & ECCV'26 & N & 25.50 & 56.50 & 69.80 & -- \\
& CIReVL & ICLR'24 & Y & 23.94 & 52.51 & 66.00 & 86.95 \\
& LDRE & SIGIR'24 & Y & 25.69 & 55.13 & 69.04 & 89.90 \\
& OSrCIR & CVPR'25 & Y & 25.42 & 54.54 & 68.19 & -- \\
& AutoCIR & KDD'25 & Y & \textbf{30.53} & \textbf{59.42} & \textbf{72.19} & \underline{91.47} \\
& ASAP-CIR (Ours) & -- & Y & \underline{28.41} & \underline{57.81} & 69.57 & 88.15 \\
\midrule
\multirow{12}{*}{ViT-L/14}
& Pic2Word & CVPR'23 & N & 23.90 & 51.70 & 65.30 & 87.80 \\
& SEARLE & ICCV'23 & N & 24.24 & 52.48 & 66.29 & 88.84 \\
& LinCIR & CVPR'24 & N & 25.08 & 53.63 & 67.30 & 88.72 \\
& DiffComp & CVPR'26 & N & \underline{32.36} & \textbf{62.90} & \textbf{74.88} & \textbf{93.23} \\
& TT-RLDR & AAAI'26 & N & 29.71 & -- & 73.02 & \underline{92.16} \\
& FlowCIR & ECCV'26 & N & 26.20 & 56.10 & 69.80 & -- \\
& CIReVL & ICLR'24 & Y & 24.55 & 52.31 & 64.92 & 86.34 \\
& LDRE & SIGIR'24 & Y & 26.53 & 55.57 & 67.54 & 88.50 \\
& AutoCIR & KDD'25 & Y & 31.81 & \underline{61.95} & \underline{73.86} & 92.07 \\
& OSrCIR & CVPR'25 & Y & 29.45 & 57.68 & 69.86 & -- \\
& STiTch & CVPR'26 & Y & 28.87 & 57.97 & 69.90 & -- \\
& ASAP-CIR (Ours) & -- & Y & \textbf{32.80} & 59.69 & 70.87 & 89.16 \\
\bottomrule
\end{tabular}
\endgroup
\end{table*}

\begin{table*}[t]
\centering
\caption{CIRCO test results. The best and second-best results within each backbone group are shown in bold and underlined, respectively.}
\label{tab:circo_results}
\begingroup
\scriptsize
\setlength{\tabcolsep}{5pt}
\renewcommand{\arraystretch}{1.08}
\begin{tabular}{@{}llcccccc@{}}
\toprule
Backbone & Method & Venue & T-Free & mAP@5 & mAP@10 & mAP@25 & mAP@50 \\
\midrule
\multirow{11}{*}{ViT-L/14}
& Pic2Word & CVPR'23 & N & 8.72 & 9.51 & 10.64 & 11.29 \\
& SEARLE & ICCV'23 & N & 11.68 & 12.73 & 14.33 & 15.12 \\
& LinCIR & CVPR'24 & N & 12.62 & 13.40 & 14.81 & 15.69 \\
& DiffComp & CVPR'26 & N & 16.19 & 17.32 & 19.24 & 20.30 \\
& FlowCIR & ECCV'26 & N & 14.90 & 15.70 & 17.30 & 18.20 \\
& CIReVL & ICLR'24 & Y & 18.57 & 19.01 & 20.89 & 21.80 \\
& LDRE & SIGIR'24 & Y & 23.35 & 24.03 & 26.44 & 27.50 \\
& AutoCIR & KDD'25 & Y & 24.05 & 25.14 & 27.35 & 28.36 \\
& OSrCIR & CVPR'25 & Y & 23.87 & 25.33 & 27.84 & 28.97 \\
& STiTch & CVPR'26 & Y & \underline{25.55} & \underline{26.27} & \underline{28.81} & \underline{29.99} \\
& ASAP-CIR (Ours) & -- & Y & \textbf{35.49} & \textbf{36.46} & \textbf{39.21} & \textbf{40.35} \\
\midrule
\multirow{6}{*}{ViT-G/14}
& LinCIR & CVPR'24 & N & 19.71 & 21.01 & 23.13 & 24.18 \\
& CIReVL & ICLR'24 & Y & 26.77 & 27.59 & 29.96 & 31.03 \\
& LDRE & SIGIR'24 & Y & 31.12 & 32.24 & 34.95 & 36.03 \\
& OSrCIR & CVPR'25 & Y & 30.47 & 31.14 & 35.03 & 36.59 \\
& STiTch & CVPR'26 & Y & \underline{34.40} & \underline{35.56} & \underline{38.07} & \underline{40.02} \\
& ASAP-CIR (Ours) & -- & Y & \textbf{37.06} & \textbf{38.47} & \textbf{41.44} & \textbf{42.52} \\
\bottomrule
\end{tabular}
\endgroup
\end{table*}

%% file: main.bbl

\begin{thebibliography}{42}
\ifx \bisbn   \undefined \def \bisbn  #1{ISBN #1}\fi
\ifx \binits  \undefined \def \binits#1{#1}\fi
\ifx \bauthor  \undefined \def \bauthor#1{#1}\fi
\ifx \batitle  \undefined \def \batitle#1{#1}\fi
\ifx \bjtitle  \undefined \def \bjtitle#1{#1}\fi
\ifx \bvolume  \undefined \def \bvolume#1{\textbf{#1}}\fi
\ifx \byear  \undefined \def \byear#1{#1}\fi
\ifx \bissue  \undefined \def \bissue#1{#1}\fi
\ifx \bfpage  \undefined \def \bfpage#1{#1}\fi
\ifx \blpage  \undefined \def \blpage #1{#1}\fi
\ifx \burl  \undefined \def \burl#1{\textsf{#1}}\fi
\ifx \doiurl  \undefined \def \doiurl#1{\url{https://doi.org/#1}}\fi
\ifx \betal  \undefined \def \betal{\textit{et al.}}\fi
\ifx \binstitute  \undefined \def \binstitute#1{#1}\fi
\ifx \binstitutionaled  \undefined \def \binstitutionaled#1{#1}\fi
\ifx \bctitle  \undefined \def \bctitle#1{#1}\fi
\ifx \beditor  \undefined \def \beditor#1{#1}\fi
\ifx \bpublisher  \undefined \def \bpublisher#1{#1}\fi
\ifx \bbtitle  \undefined \def \bbtitle#1{#1}\fi
\ifx \bedition  \undefined \def \bedition#1{#1}\fi
\ifx \bseriesno  \undefined \def \bseriesno#1{#1}\fi
\ifx \blocation  \undefined \def \blocation#1{#1}\fi
\ifx \bsertitle  \undefined \def \bsertitle#1{#1}\fi
\ifx \bsnm \undefined \def \bsnm#1{#1}\fi
\ifx \bsuffix \undefined \def \bsuffix#1{#1}\fi
\ifx \bparticle \undefined \def \bparticle#1{#1}\fi
\ifx \barticle \undefined \def \barticle#1{#1}\fi
\bibcommenthead
\ifx \bconfdate \undefined \def \bconfdate #1{#1}\fi
\ifx \botherref \undefined \def \botherref #1{#1}\fi
\ifx \url \undefined \def \url#1{\textsf{#1}}\fi
\ifx \bchapter \undefined \def \bchapter#1{#1}\fi
\ifx \bbook \undefined \def \bbook#1{#1}\fi
\ifx \bcomment \undefined \def \bcomment#1{#1}\fi
\ifx \oauthor \undefined \def \oauthor#1{#1}\fi
\ifx \citeauthoryear \undefined \def \citeauthoryear#1{#1}\fi
\ifx \endbibitem  \undefined \def \endbibitem {}\fi
\ifx \bconflocation  \undefined \def \bconflocation#1{#1}\fi
\ifx \arxivurl  \undefined \def \arxivurl#1{\textsf{#1}}\fi
\csname PreBibitemsHook\endcsname

%%% 1
\bibitem[\protect\citeauthoryear{An et~al.}{2024}]{ICLR2024_8e5b9dc3}
\begin{bchapter}
\bauthor{\bsnm{An}, \binits{B.}},
\bauthor{\bsnm{Zhu}, \binits{S.}},
\bauthor{\bsnm{Panaitescu-Liess}, \binits{M.-A.}},
\bauthor{\bsnm{Mummadi}, \binits{C.K.}},
\bauthor{\bsnm{Huang}, \binits{F.}}:
\bctitle{Perceptionclip: Visual classification by inferring and conditioning on contexts}.
In: \beditor{\bsnm{Kim}, \binits{B.}},
\beditor{\bsnm{Yue}, \binits{Y.}},
\beditor{\bsnm{Chaudhuri}, \binits{S.}},
\beditor{\bsnm{Fragkiadaki}, \binits{K.}},
\beditor{\bsnm{Khan}, \binits{M.}},
\beditor{\bsnm{Sun}, \binits{Y.}} (eds.)
\bbtitle{International Conference on Learning Representations},
vol. \bseriesno{2024},
pp. \bfpage{33373}--\blpage{33398}
(\byear{2024})
\end{bchapter}
\endbibitem

%%% 2
\bibitem[\protect\citeauthoryear{Baldrati et~al.}{2023}]{Baldrati_2023_ICCV}
\begin{bchapter}
\bauthor{\bsnm{Baldrati}, \binits{A.}},
\bauthor{\bsnm{Agnolucci}, \binits{L.}},
\bauthor{\bsnm{Bertini}, \binits{M.}},
\bauthor{\bsnm{Del~Bimbo}, \binits{A.}}:
\bctitle{Zero-shot composed image retrieval with textual inversion}.
In: \bbtitle{Proceedings of the IEEE/CVF International Conference on Computer Vision (ICCV)},
pp. \bfpage{15338}--\blpage{15347}
(\byear{2023})
\end{bchapter}
\endbibitem

%%% 3
\bibitem[\protect\citeauthoryear{Baldrati et~al.}{2022}]{Baldrati_2022_CVPR}
\begin{bchapter}
\bauthor{\bsnm{Baldrati}, \binits{A.}},
\bauthor{\bsnm{Bertini}, \binits{M.}},
\bauthor{\bsnm{Uricchio}, \binits{T.}},
\bauthor{\bsnm{Del~Bimbo}, \binits{A.}}:
\bctitle{Effective conditioned and composed image retrieval combining clip-based features}.
In: \bbtitle{Proceedings of the IEEE/CVF Conference on Computer Vision and Pattern Recognition (CVPR)},
pp. \bfpage{21466}--\blpage{21474}
(\byear{2022})
\end{bchapter}
\endbibitem

%%% 4
\bibitem[\protect\citeauthoryear{Cao et~al.}{2026}]{cao2026cocoir}
\begin{bchapter}
\bauthor{\bsnm{Cao}, \binits{S.}},
\bauthor{\bsnm{Dabral}, \binits{T.S.}},
\bauthor{\bsnm{Ding}, \binits{Z.}},
\bauthor{\bsnm{Shanbhogue}, \binits{M.}},
\bauthor{\bsnm{Chen}, \binits{K.}},
\bauthor{\bsnm{Li}, \binits{Z.}},
\bauthor{\bsnm{Seyedhosseini}, \binits{M.}},
\bauthor{\bsnm{Gui}, \binits{L.-Y.}},
\bauthor{\bsnm{Wang}, \binits{Y.-X.}}:
\bctitle{Coco-ir: Contextual composed image retrieval}.
In: \bbtitle{ECCV}
(\byear{2026})
\end{bchapter}
\endbibitem

%%% 5
\bibitem[\protect\citeauthoryear{Cheng et~al.}{2025}]{10.1145/3711896.3736982}
\begin{bchapter}
\bauthor{\bsnm{Cheng}, \binits{Z.}},
\bauthor{\bsnm{Ma}, \binits{Y.}},
\bauthor{\bsnm{Lang}, \binits{J.}},
\bauthor{\bsnm{Zhang}, \binits{K.}},
\bauthor{\bsnm{Zhong}, \binits{T.}},
\bauthor{\bsnm{Wang}, \binits{Y.}},
\bauthor{\bsnm{Zhou}, \binits{F.}}:
\bctitle{Generative thinking, corrective action: User-friendly composed image retrieval via automatic multi-agent collaboration}.
In: \bbtitle{Proceedings of the 31st ACM SIGKDD Conference on Knowledge Discovery and Data Mining V.2}.
\bsertitle{KDD '25},
pp. \bfpage{334}--\blpage{344}.
\bpublisher{Association for Computing Machinery},
\blocation{New York, NY, USA}
(\byear{2025}).
\doiurl{10.1145/3711896.3736982} .
\burl{https://doi.org/10.1145/3711896.3736982}
\end{bchapter}
\endbibitem

%%% 6
\bibitem[\protect\citeauthoryear{Chen and Wu}{2024}]{Chen_2024_CVPR}
\begin{bchapter}
\bauthor{\bsnm{Chen}, \binits{K.}},
\bauthor{\bsnm{Wu}, \binits{X.}}:
\bctitle{Vtqa: Visual text question answering via entity alignment and cross-media reasoning}.
In: \bbtitle{Proceedings of the IEEE/CVF Conference on Computer Vision and Pattern Recognition (CVPR)},
pp. \bfpage{27218}--\blpage{27227}
(\byear{2024})
\end{bchapter}
\endbibitem

%%% 7
\bibitem[\protect\citeauthoryear{Chen et~al.}{2026}]{CHEN2026113187}
\begin{barticle}
\bauthor{\bsnm{Chen}, \binits{Z.}},
\bauthor{\bsnm{Zhao}, \binits{Z.}},
\bauthor{\bsnm{Su}, \binits{F.}},
\bauthor{\bsnm{Lu}, \binits{S.}}:
\batitle{Data-efficient generalization for zero-shot composed image retrieval}.
\bjtitle{Pattern Recognition}
\bvolume{176},
\bfpage{113187}
(\byear{2026})
\doiurl{10.1016/j.patcog.2026.113187}
\end{barticle}
\endbibitem

%%% 8
\bibitem[\protect\citeauthoryear{Fu et~al.}{2026}]{Fu_2026_CVPR}
\begin{bchapter}
\bauthor{\bsnm{Fu}, \binits{Z.}},
\bauthor{\bsnm{Hu}, \binits{Y.}},
\bauthor{\bsnm{Yang}, \binits{Q.}},
\bauthor{\bsnm{Zhang}, \binits{S.}},
\bauthor{\bsnm{Chen}, \binits{Z.}},
\bauthor{\bsnm{Li}, \binits{Z.}}:
\bctitle{Air-know: Arbiter-calibrated knowledge-internalizing robust network for composed image retrieval}.
In: \bbtitle{Proceedings of the IEEE/CVF Conference on Computer Vision and Pattern Recognition (CVPR)},
pp. \bfpage{2658}--\blpage{2670}
(\byear{2026})
\end{bchapter}
\endbibitem

%%% 9
\bibitem[\protect\citeauthoryear{Gu et~al.}{2024}]{Gu_2024_CVPR}
\begin{bchapter}
\bauthor{\bsnm{Gu}, \binits{G.}},
\bauthor{\bsnm{Chun}, \binits{S.}},
\bauthor{\bsnm{Kim}, \binits{W.}},
\bauthor{\bsnm{Kang}, \binits{Y.}},
\bauthor{\bsnm{Yun}, \binits{S.}}:
\bctitle{Language-only training of zero-shot composed image retrieval}.
In: \bbtitle{Proceedings of the IEEE/CVF Conference on Computer Vision and Pattern Recognition (CVPR)},
pp. \bfpage{13225}--\blpage{13234}
(\byear{2024})
\end{bchapter}
\endbibitem

%%% 10
\bibitem[\protect\citeauthoryear{Guo et~al.}{2023}]{Guo_2023_CVPR}
\begin{bchapter}
\bauthor{\bsnm{Guo}, \binits{J.}},
\bauthor{\bsnm{Li}, \binits{J.}},
\bauthor{\bsnm{Li}, \binits{D.}},
\bauthor{\bsnm{Tiong}, \binits{A.M.H.}},
\bauthor{\bsnm{Li}, \binits{B.}},
\bauthor{\bsnm{Tao}, \binits{D.}},
\bauthor{\bsnm{Hoi}, \binits{S.}}:
\bctitle{From images to textual prompts: Zero-shot visual question answering with frozen large language models}.
In: \bbtitle{Proceedings of the IEEE/CVF Conference on Computer Vision and Pattern Recognition (CVPR)},
pp. \bfpage{10867}--\blpage{10877}
(\byear{2023})
\end{bchapter}
\endbibitem

%%% 11
\bibitem[\protect\citeauthoryear{Ge et~al.}{2023}]{Ge_2023_CVPR}
\begin{bchapter}
\bauthor{\bsnm{Ge}, \binits{Y.}},
\bauthor{\bsnm{Ren}, \binits{J.}},
\bauthor{\bsnm{Gallagher}, \binits{A.}},
\bauthor{\bsnm{Wang}, \binits{Y.}},
\bauthor{\bsnm{Yang}, \binits{M.-H.}},
\bauthor{\bsnm{Adam}, \binits{H.}},
\bauthor{\bsnm{Itti}, \binits{L.}},
\bauthor{\bsnm{Lakshminarayanan}, \binits{B.}},
\bauthor{\bsnm{Zhao}, \binits{J.}}:
\bctitle{Improving zero-shot generalization and robustness of multi-modal models}.
In: \bbtitle{Proceedings of the IEEE/CVF Conference on Computer Vision and Pattern Recognition (CVPR)},
pp. \bfpage{11093}--\blpage{11101}
(\byear{2023})
\end{bchapter}
\endbibitem

%%% 12
\bibitem[\protect\citeauthoryear{Huang et~al.}{2024}]{Huang_2024_CVPR}
\begin{bchapter}
\bauthor{\bsnm{Huang}, \binits{S.}},
\bauthor{\bsnm{Gong}, \binits{B.}},
\bauthor{\bsnm{Feng}, \binits{Y.}},
\bauthor{\bsnm{Zhang}, \binits{M.}},
\bauthor{\bsnm{Lv}, \binits{Y.}},
\bauthor{\bsnm{Wang}, \binits{D.}}:
\bctitle{Troika: Multi-path cross-modal traction for compositional zero-shot learning}.
In: \bbtitle{Proceedings of the IEEE/CVF Conference on Computer Vision and Pattern Recognition (CVPR)},
pp. \bfpage{24005}--\blpage{24014}
(\byear{2024})
\end{bchapter}
\endbibitem

%%% 13
\bibitem[\protect\citeauthoryear{He et~al.}{2026}]{he2026flowcir}
\begin{bchapter}
\bauthor{\bsnm{He}, \binits{Z.}},
\bauthor{\bsnm{Jiang}, \binits{Z.}},
\bauthor{\bsnm{Liu}, \binits{Y.}},
\bauthor{\bsnm{Wang}, \binits{Y.}},
\bauthor{\bsnm{Wang}, \binits{T.}},
\bauthor{\bsnm{Chen}, \binits{L.}}:
\bctitle{Flowcir: Semantic transport via flow matching for zero-shot composed image retrieval}.
In: \bbtitle{Computer Vision -- ECCV 2026},
pp. \bfpage{163}--\blpage{181}.
\bpublisher{Springer},
\blocation{Cham}
(\byear{2026})
\end{bchapter}
\endbibitem

%%% 14
\bibitem[\protect\citeauthoryear{Han et~al.}{2017}]{Han_2017_ICCV}
\begin{bchapter}
\bauthor{\bsnm{Han}, \binits{X.}},
\bauthor{\bsnm{Wu}, \binits{Z.}},
\bauthor{\bsnm{Huang}, \binits{P.X.}},
\bauthor{\bsnm{Zhang}, \binits{X.}},
\bauthor{\bsnm{Zhu}, \binits{M.}},
\bauthor{\bsnm{Li}, \binits{Y.}},
\bauthor{\bsnm{Zhao}, \binits{Y.}},
\bauthor{\bsnm{Davis}, \binits{L.S.}}:
\bctitle{Automatic spatially-aware fashion concept discovery}.
In: \bbtitle{Proceedings of the IEEE International Conference on Computer Vision (ICCV)}
(\byear{2017})
\end{bchapter}
\endbibitem

%%% 15
\bibitem[\protect\citeauthoryear{Han et~al.}{2023}]{Han_2023_CVPR}
\begin{bchapter}
\bauthor{\bsnm{Han}, \binits{X.}},
\bauthor{\bsnm{Zhu}, \binits{X.}},
\bauthor{\bsnm{Yu}, \binits{L.}},
\bauthor{\bsnm{Zhang}, \binits{L.}},
\bauthor{\bsnm{Song}, \binits{Y.-Z.}},
\bauthor{\bsnm{Xiang}, \binits{T.}}:
\bctitle{Fame-vil: Multi-tasking vision-language model for heterogeneous fashion tasks}.
In: \bbtitle{Proceedings of the IEEE/CVF Conference on Computer Vision and Pattern Recognition (CVPR)},
pp. \bfpage{2669}--\blpage{2680}
(\byear{2023})
\end{bchapter}
\endbibitem

%%% 16
\bibitem[\protect\citeauthoryear{Jeong et~al.}{2023}]{Jeong_2023_CVPR}
\begin{bchapter}
\bauthor{\bsnm{Jeong}, \binits{J.}},
\bauthor{\bsnm{Zou}, \binits{Y.}},
\bauthor{\bsnm{Kim}, \binits{T.}},
\bauthor{\bsnm{Zhang}, \binits{D.}},
\bauthor{\bsnm{Ravichandran}, \binits{A.}},
\bauthor{\bsnm{Dabeer}, \binits{O.}}:
\bctitle{Winclip: Zero-/few-shot anomaly classification and segmentation}.
In: \bbtitle{Proceedings of the IEEE/CVF Conference on Computer Vision and Pattern Recognition (CVPR)},
pp. \bfpage{19606}--\blpage{19616}
(\byear{2023})
\end{bchapter}
\endbibitem

%%% 17
\bibitem[\protect\citeauthoryear{Karthik et~al.}{2024}]{karthik2024visionbylanguage}
\begin{bchapter}
\bauthor{\bsnm{Karthik}, \binits{S.}},
\bauthor{\bsnm{Roth}, \binits{K.}},
\bauthor{\bsnm{Mancini}, \binits{M.}},
\bauthor{\bsnm{Akata}, \binits{Z.}}:
\bctitle{Vision-by-language for training-free compositional image retrieval}.
In: \bbtitle{The Twelfth International Conference on Learning Representations}
(\byear{2024})
\end{bchapter}
\endbibitem

%%% 18
\bibitem[\protect\citeauthoryear{Lu et~al.}{2023}]{Lu_2023_CVPR}
\begin{bchapter}
\bauthor{\bsnm{Lu}, \binits{X.}},
\bauthor{\bsnm{Guo}, \binits{S.}},
\bauthor{\bsnm{Liu}, \binits{Z.}},
\bauthor{\bsnm{Guo}, \binits{J.}}:
\bctitle{Decomposed soft prompt guided fusion enhancing for compositional zero-shot learning}.
In: \bbtitle{Proceedings of the IEEE/CVF Conference on Computer Vision and Pattern Recognition (CVPR)},
pp. \bfpage{23560}--\blpage{23569}
(\byear{2023})
\end{bchapter}
\endbibitem

%%% 19
\bibitem[\protect\citeauthoryear{Liu et~al.}{2021}]{Liu_2021_ICCV}
\begin{bchapter}
\bauthor{\bsnm{Liu}, \binits{Z.}},
\bauthor{\bsnm{Rodriguez-Opazo}, \binits{C.}},
\bauthor{\bsnm{Teney}, \binits{D.}},
\bauthor{\bsnm{Gould}, \binits{S.}}:
\bctitle{Image retrieval on real-life images with pre-trained vision-and-language models}.
In: \bbtitle{Proceedings of the IEEE/CVF International Conference on Computer Vision (ICCV)},
pp. \bfpage{2125}--\blpage{2134}
(\byear{2021})
\end{bchapter}
\endbibitem

%%% 20
\bibitem[\protect\citeauthoryear{Li et~al.}{2026}]{Li_2026_CVPR}
\begin{bchapter}
\bauthor{\bsnm{Li}, \binits{M.}},
\bauthor{\bsnm{Wang}, \binits{D.}},
\bauthor{\bsnm{Sun}, \binits{Z.}},
\bauthor{\bsnm{Zhang}, \binits{J.}},
\bauthor{\bsnm{Luo}, \binits{W.}},
\bauthor{\bsnm{Guo}, \binits{J.}}:
\bctitle{Stitch: Semantic transition and transportation in collaboration for training-free zero-shot composed image retrieval}.
In: \bbtitle{Proceedings of the IEEE/CVF Conference on Computer Vision and Pattern Recognition (CVPR)},
pp. \bfpage{12374}--\blpage{12384}
(\byear{2026})
\end{bchapter}
\endbibitem

%%% 21
\bibitem[\protect\citeauthoryear{Lim et~al.}{2026}]{lim2026gmixer}
\begin{bchapter}
\bauthor{\bsnm{Lim}, \binits{J.}},
\bauthor{\bsnm{Yang}, \binits{H.}},
\bauthor{\bsnm{Lee}, \binits{J.-H.}}:
\bctitle{G-mixer: Geodesic mixup-based implicit semantic expansion and explicit semantic re-ranking for zero-shot composed image retrieval}.
In: \bbtitle{Proceedings of the IEEE/CVF Conference on Computer Vision and Pattern Recognition (CVPR)},
pp. \bfpage{33891}--\blpage{33900}
(\byear{2026})
\end{bchapter}
\endbibitem

%%% 22
\bibitem[\protect\citeauthoryear{Li et~al.}{2025}]{10.1145/3726302.3730074}
\begin{bchapter}
\bauthor{\bsnm{Li}, \binits{Z.}},
\bauthor{\bsnm{Zhang}, \binits{L.}},
\bauthor{\bsnm{Zhang}, \binits{K.}},
\bauthor{\bsnm{Chen}, \binits{W.}},
\bauthor{\bsnm{Zhang}, \binits{Y.}},
\bauthor{\bsnm{Mao}, \binits{Z.}}:
\bctitle{Rethinking pseudo word learning in zero-shot composed image retrieval: From an object-aware perspective}.
In: \bbtitle{Proceedings of the 48th International ACM SIGIR Conference on Research and Development in Information Retrieval}.
\bsertitle{SIGIR '25},
pp. \bfpage{833}--\blpage{843}.
\bpublisher{Association for Computing Machinery},
\blocation{New York, NY, USA}
(\byear{2025}).
\doiurl{10.1145/3726302.3730074} .
\burl{https://doi.org/10.1145/3726302.3730074}
\end{bchapter}
\endbibitem

%%% 23
\bibitem[\protect\citeauthoryear{Martin et~al.}{2024}]{Martin_2024_CVPR}
\begin{bchapter}
\bauthor{\bsnm{Martin}, \binits{S.}},
\bauthor{\bsnm{Huang}, \binits{Y.}},
\bauthor{\bsnm{Shakeri}, \binits{F.}},
\bauthor{\bsnm{Pesquet}, \binits{J.-C.}},
\bauthor{\bsnm{Ben~Ayed}, \binits{I.}}:
\bctitle{Transductive zero-shot and few-shot clip}.
In: \bbtitle{Proceedings of the IEEE/CVF Conference on Computer Vision and Pattern Recognition (CVPR)},
pp. \bfpage{28816}--\blpage{28826}
(\byear{2024})
\end{bchapter}
\endbibitem

%%% 24
\bibitem[\protect\citeauthoryear{Radford et~al.}{2021}]{pmlr-v139-radford21a}
\begin{bchapter}
\bauthor{\bsnm{Radford}, \binits{A.}},
\bauthor{\bsnm{Kim}, \binits{J.W.}},
\bauthor{\bsnm{Hallacy}, \binits{C.}},
\bauthor{\bsnm{Ramesh}, \binits{A.}},
\bauthor{\bsnm{Goh}, \binits{G.}},
\bauthor{\bsnm{Agarwal}, \binits{S.}},
\bauthor{\bsnm{Sastry}, \binits{G.}},
\bauthor{\bsnm{Askell}, \binits{A.}},
\bauthor{\bsnm{Mishkin}, \binits{P.}},
\bauthor{\bsnm{Clark}, \binits{J.}},
\bauthor{\bsnm{Krueger}, \binits{G.}},
\bauthor{\bsnm{Sutskever}, \binits{I.}}:
\bctitle{Learning transferable visual models from natural language supervision}.
In: \beditor{\bsnm{Meila}, \binits{M.}},
\beditor{\bsnm{Zhang}, \binits{T.}} (eds.)
\bbtitle{Proceedings of the 38th International Conference on Machine Learning}.
\bsertitle{Proceedings of Machine Learning Research},
vol. \bseriesno{139},
pp. \bfpage{8748}--\blpage{8763}.
\bpublisher{PMLR},
\blocation{Online}
(\byear{2021})
\end{bchapter}
\endbibitem

%%% 25
\bibitem[\protect\citeauthoryear{Song et~al.}{2024}]{Song_2024_CVPR}
\begin{bchapter}
\bauthor{\bsnm{Song}, \binits{C.H.}},
\bauthor{\bsnm{Hwang}, \binits{T.}},
\bauthor{\bsnm{Yoon}, \binits{J.}},
\bauthor{\bsnm{Choi}, \binits{S.}},
\bauthor{\bsnm{Gu}, \binits{Y.H.}}:
\bctitle{Syncmask: Synchronized attentional masking for fashion-centric vision-language pretraining}.
In: \bbtitle{Proceedings of the IEEE/CVF Conference on Computer Vision and Pattern Recognition (CVPR)},
pp. \bfpage{13948}--\blpage{13957}
(\byear{2024})
\end{bchapter}
\endbibitem

%%% 26
\bibitem[\protect\citeauthoryear{Sun et~al.}{2025}]{Sun_2025_ICCV}
\begin{bchapter}
\bauthor{\bsnm{Sun}, \binits{Z.}},
\bauthor{\bsnm{Jing}, \binits{D.}},
\bauthor{\bsnm{Lu}, \binits{Z.}}:
\bctitle{Cotmr: Chain-of-thought multi-scale reasoning for training-free zero-shot composed image retrieval}.
In: \bbtitle{Proceedings of the IEEE/CVF International Conference on Computer Vision (ICCV)},
pp. \bfpage{22675}--\blpage{22684}
(\byear{2025})
\end{bchapter}
\endbibitem

%%% 27
\bibitem[\protect\citeauthoryear{Saito et~al.}{2023}]{Saito_2023_CVPR}
\begin{bchapter}
\bauthor{\bsnm{Saito}, \binits{K.}},
\bauthor{\bsnm{Sohn}, \binits{K.}},
\bauthor{\bsnm{Zhang}, \binits{X.}},
\bauthor{\bsnm{Li}, \binits{C.-L.}},
\bauthor{\bsnm{Lee}, \binits{C.-Y.}},
\bauthor{\bsnm{Saenko}, \binits{K.}},
\bauthor{\bsnm{Pfister}, \binits{T.}}:
\bctitle{Pic2word: Mapping pictures to words for zero-shot composed image retrieval}.
In: \bbtitle{Proceedings of the IEEE/CVF Conference on Computer Vision and Pattern Recognition (CVPR)},
pp. \bfpage{19305}--\blpage{19314}
(\byear{2023})
\end{bchapter}
\endbibitem

%%% 28
\bibitem[\protect\citeauthoryear{Saha et~al.}{2024}]{Saha_2024_CVPR}
\begin{bchapter}
\bauthor{\bsnm{Saha}, \binits{O.}},
\bauthor{\bsnm{Van~Horn}, \binits{G.}},
\bauthor{\bsnm{Maji}, \binits{S.}}:
\bctitle{Improved zero-shot classification by adapting vlms with text descriptions}.
In: \bbtitle{Proceedings of the IEEE/CVF Conference on Computer Vision and Pattern Recognition (CVPR)},
pp. \bfpage{17542}--\blpage{17552}
(\byear{2024})
\end{bchapter}
\endbibitem

%%% 29
\bibitem[\protect\citeauthoryear{Tschannen et~al.}{2023}]{Tschannen_2023_CVPR}
\begin{bchapter}
\bauthor{\bsnm{Tschannen}, \binits{M.}},
\bauthor{\bsnm{Mustafa}, \binits{B.}},
\bauthor{\bsnm{Houlsby}, \binits{N.}}:
\bctitle{Clippo: Image-and-language understanding from pixels only}.
In: \bbtitle{Proceedings of the IEEE/CVF Conference on Computer Vision and Pattern Recognition (CVPR)},
pp. \bfpage{11006}--\blpage{11017}
(\byear{2023})
\end{bchapter}
\endbibitem

%%% 30
\bibitem[\protect\citeauthoryear{Tang et~al.}{2025}]{Tang_2025_PrediCIR}
\begin{bchapter}
\bauthor{\bsnm{Tang}, \binits{Y.}},
\bauthor{\bsnm{Yu}, \binits{J.}},
\bauthor{\bsnm{Gai}, \binits{K.}},
\bauthor{\bsnm{Zhuang}, \binits{J.}},
\bauthor{\bsnm{Xiong}, \binits{G.}},
\bauthor{\bsnm{Gou}, \binits{G.}},
\bauthor{\bsnm{Wu}, \binits{Q.}}:
\bctitle{Missing target-relevant information prediction with world model for accurate zero-shot composed image retrieval}.
In: \bbtitle{Proceedings of the IEEE/CVF Conference on Computer Vision and Pattern Recognition (CVPR)},
pp. \bfpage{24785}--\blpage{24795}
(\byear{2025})
\end{bchapter}
\endbibitem

%%% 31
\bibitem[\protect\citeauthoryear{Tang et~al.}{2025}]{Tang_2025_OSrCIR}
\begin{bchapter}
\bauthor{\bsnm{Tang}, \binits{Y.}},
\bauthor{\bsnm{Zhang}, \binits{J.}},
\bauthor{\bsnm{Qin}, \binits{X.}},
\bauthor{\bsnm{Yu}, \binits{J.}},
\bauthor{\bsnm{Gou}, \binits{G.}},
\bauthor{\bsnm{Xiong}, \binits{G.}},
\bauthor{\bsnm{Lin}, \binits{Q.}},
\bauthor{\bsnm{Rajmohan}, \binits{S.}},
\bauthor{\bsnm{Zhang}, \binits{D.}},
\bauthor{\bsnm{Wu}, \binits{Q.}}:
\bctitle{Reason-before-retrieve: One-stage reflective chain-of-thoughts for training-free zero-shot composed image retrieval}.
In: \bbtitle{Proceedings of the IEEE/CVF Conference on Computer Vision and Pattern Recognition (CVPR)},
pp. \bfpage{14400}--\blpage{14410}
(\byear{2025})
\end{bchapter}
\endbibitem

%%% 32
\bibitem[\protect\citeauthoryear{Vo et~al.}{2019}]{Vo_2019_CVPR}
\begin{bchapter}
\bauthor{\bsnm{Vo}, \binits{N.}},
\bauthor{\bsnm{Jiang}, \binits{L.}},
\bauthor{\bsnm{Sun}, \binits{C.}},
\bauthor{\bsnm{Murphy}, \binits{K.}},
\bauthor{\bsnm{Li}, \binits{L.-J.}},
\bauthor{\bsnm{Fei-Fei}, \binits{L.}},
\bauthor{\bsnm{Hays}, \binits{J.}}:
\bctitle{Composing text and image for image retrieval - an empirical odyssey}.
In: \bbtitle{Proceedings of the IEEE/CVF Conference on Computer Vision and Pattern Recognition (CVPR)}
(\byear{2019})
\end{bchapter}
\endbibitem

%%% 33
\bibitem[\protect\citeauthoryear{Wu et~al.}{2021}]{Wu_2021_CVPR}
\begin{bchapter}
\bauthor{\bsnm{Wu}, \binits{H.}},
\bauthor{\bsnm{Gao}, \binits{Y.}},
\bauthor{\bsnm{Guo}, \binits{X.}},
\bauthor{\bsnm{Al-Halah}, \binits{Z.}},
\bauthor{\bsnm{Rennie}, \binits{S.}},
\bauthor{\bsnm{Grauman}, \binits{K.}},
\bauthor{\bsnm{Feris}, \binits{R.}}:
\bctitle{Fashion iq: A new dataset towards retrieving images by natural language feedback}.
In: \bbtitle{Proceedings of the IEEE/CVF Conference on Computer Vision and Pattern Recognition (CVPR)},
pp. \bfpage{11307}--\blpage{11317}
(\byear{2021})
\end{bchapter}
\endbibitem

%%% 34
\bibitem[\protect\citeauthoryear{Wu et~al.}{2025}]{Wu_2025_CVPR}
\begin{bchapter}
\bauthor{\bsnm{Wu}, \binits{P.}},
\bauthor{\bsnm{Lu}, \binits{X.}},
\bauthor{\bsnm{Hu}, \binits{H.}},
\bauthor{\bsnm{Xian}, \binits{Y.}},
\bauthor{\bsnm{Shen}, \binits{J.}},
\bauthor{\bsnm{Wang}, \binits{W.}}:
\bctitle{Logiczsl: Exploring logic-induced representation for compositional zero-shot learning}.
In: \bbtitle{Proceedings of the IEEE/CVF Conference on Computer Vision and Pattern Recognition (CVPR)},
pp. \bfpage{30301}--\blpage{30311}
(\byear{2025})
\end{bchapter}
\endbibitem

%%% 35
\bibitem[\protect\citeauthoryear{Wang et~al.}{2024}]{ICLR2024_7c40c505}
\begin{bchapter}
\bauthor{\bsnm{Wang}, \binits{Z.}},
\bauthor{\bsnm{Liang}, \binits{J.}},
\bauthor{\bsnm{Sheng}, \binits{L.}},
\bauthor{\bsnm{He}, \binits{R.}},
\bauthor{\bsnm{Wang}, \binits{Z.}},
\bauthor{\bsnm{Tan}, \binits{T.}}:
\bctitle{A hard-to-beat baseline for training-free clip-based adaptation}.
In: \beditor{\bsnm{Kim}, \binits{B.}},
\beditor{\bsnm{Yue}, \binits{Y.}},
\beditor{\bsnm{Chaudhuri}, \binits{S.}},
\beditor{\bsnm{Fragkiadaki}, \binits{K.}},
\beditor{\bsnm{Khan}, \binits{M.}},
\beditor{\bsnm{Sun}, \binits{Y.}} (eds.)
\bbtitle{International Conference on Learning Representations},
vol. \bseriesno{2024},
pp. \bfpage{28841}--\blpage{28859}
(\byear{2024})
\end{bchapter}
\endbibitem

%%% 36
\bibitem[\protect\citeauthoryear{Wang et~al.}{2026}]{wang2026wiser}
\begin{bchapter}
\bauthor{\bsnm{Wang}, \binits{T.}},
\bauthor{\bsnm{Qu}, \binits{L.}},
\bauthor{\bsnm{Yang}, \binits{T.}},
\bauthor{\bsnm{Hao}, \binits{X.}},
\bauthor{\bsnm{Xu}, \binits{Y.}},
\bauthor{\bsnm{Guo}, \binits{H.}},
\bauthor{\bsnm{Wang}, \binits{J.}}:
\bctitle{Wiser: Wider search, deeper thinking, and adaptive fusion for training-free zero-shot composed image retrieval}.
In: \bbtitle{Proceedings of the IEEE/CVF Conference on Computer Vision and Pattern Recognition (CVPR)},
pp. \bfpage{16864}--\blpage{16875}
(\byear{2026})
\end{bchapter}
\endbibitem

%%% 37
\bibitem[\protect\citeauthoryear{Xie et~al.}{2026}]{ICLR2026_cb825b4e}
\begin{bchapter}
\bauthor{\bsnm{Xie}, \binits{M.-K.}},
\bauthor{\bsnm{Kou}, \binits{Z.}},
\bauthor{\bsnm{Li}, \binits{Z.}},
\bauthor{\bsnm{Niu}, \binits{G.}},
\bauthor{\bsnm{Sugiyama}, \binits{M.}}:
\bctitle{Unlocking the power of co-occurrence in clip: A dualprompt-driven method for training-free zero-shot multi-label classification}.
In: \beditor{\bsnm{Vondrick}, \binits{C.}},
\beditor{\bsnm{Hariharan}, \binits{B.}},
\beditor{\bsnm{Raffel}, \binits{C.}},
\beditor{\bsnm{Pinto}, \binits{L.}},
\beditor{\bsnm{Yang}, \binits{D.}},
\beditor{\bsnm{Faust}, \binits{A.}} (eds.)
\bbtitle{International Conference on Learning Representations},
vol. \bseriesno{2026},
pp. \bfpage{125344}--\blpage{125357}
(\byear{2026})
\end{bchapter}
\endbibitem

%%% 38
\bibitem[\protect\citeauthoryear{Yang et~al.}{2024}]{10.1145/3626772.3657740}
\begin{bchapter}
\bauthor{\bsnm{Yang}, \binits{Z.}},
\bauthor{\bsnm{Xue}, \binits{D.}},
\bauthor{\bsnm{Qian}, \binits{S.}},
\bauthor{\bsnm{Dong}, \binits{W.}},
\bauthor{\bsnm{Xu}, \binits{C.}}:
\bctitle{Ldre: Llm-based divergent reasoning and ensemble for zero-shot composed image retrieval}.
In: \bbtitle{Proceedings of the 47th International ACM SIGIR Conference on Research and Development in Information Retrieval}.
\bsertitle{SIGIR '24},
pp. \bfpage{800}--\blpage{810}.
\bpublisher{Association for Computing Machinery},
\blocation{New York, NY, USA}
(\byear{2024}).
\doiurl{10.1145/3626772.3657740} .
\burl{https://doi.org/10.1145/3626772.3657740}
\end{bchapter}
\endbibitem

%%% 39
\bibitem[\protect\citeauthoryear{Zanella and Ben~Ayed}{2024}]{Zanella_2024_CVPR}
\begin{bchapter}
\bauthor{\bsnm{Zanella}, \binits{M.}},
\bauthor{\bsnm{Ben~Ayed}, \binits{I.}}:
\bctitle{On the test-time zero-shot generalization of vision-language models: Do we really need prompt learning?}
In: \bbtitle{Proceedings of the IEEE/CVF Conference on Computer Vision and Pattern Recognition (CVPR)},
pp. \bfpage{23783}--\blpage{23793}
(\byear{2024})
\end{bchapter}
\endbibitem

%%% 40
\bibitem[\protect\citeauthoryear{Zhang et~al.}{2025}]{ICLR2025_2c2be726}
\begin{bchapter}
\bauthor{\bsnm{Zhang}, \binits{M.}},
\bauthor{\bsnm{Bi}, \binits{K.}},
\bauthor{\bsnm{Chen}, \binits{W.}},
\bauthor{\bsnm{Guo}, \binits{J.}},
\bauthor{\bsnm{Cheng}, \binits{X.}}:
\bctitle{Clipure: Purification in latent space via clip for adversarially robust zero-shot classification}.
In: \beditor{\bsnm{Yue}, \binits{Y.}},
\beditor{\bsnm{Garg}, \binits{A.}},
\beditor{\bsnm{Peng}, \binits{N.}},
\beditor{\bsnm{Sha}, \binits{F.}},
\beditor{\bsnm{Yu}, \binits{R.}} (eds.)
\bbtitle{International Conference on Learning Representations},
vol. \bseriesno{2025},
pp. \bfpage{17360}--\blpage{17384}
(\byear{2025})
\end{bchapter}
\endbibitem

%%% 41
\bibitem[\protect\citeauthoryear{Zhang et~al.}{2026}]{Zhang_2026_CVPR}
\begin{bchapter}
\bauthor{\bsnm{Zhang}, \binits{J.}},
\bauthor{\bsnm{Zhang}, \binits{L.}},
\bauthor{\bsnm{Fu}, \binits{Z.}},
\bauthor{\bsnm{Hu}, \binits{B.}},
\bauthor{\bsnm{Mao}, \binits{Z.}}:
\bctitle{Self-guided semantic inspection for zero-shot composed image retrieval}.
In: \bbtitle{Proceedings of the IEEE/CVF Conference on Computer Vision and Pattern Recognition (CVPR)},
pp. \bfpage{33881}--\blpage{33890}
(\byear{2026})
\end{bchapter}
\endbibitem

%%% 42
\bibitem[\protect\citeauthoryear{Zhou et~al.}{2026}]{Zhou2026TTRLDR}
\begin{bchapter}
\bauthor{\bsnm{Zhou}, \binits{H.}},
\bauthor{\bsnm{Zhang}, \binits{F.}},
\bauthor{\bsnm{Xu}, \binits{C.}}:
\bctitle{Duplex rewards optimization for test-time composed image retrieval}.
In: \bbtitle{Proceedings of the AAAI Conference on Artificial Intelligence},
vol. \bseriesno{40},
pp. \bfpage{13629}--\blpage{13637}
(\byear{2026}).
\doiurl{10.1609/aaai.v40i16.38369}
\end{bchapter}
\endbibitem

\end{thebibliography}
